\pdfoutput=1
\documentclass[11pt]{article}

\usepackage[preprint]{acl}

\usepackage{times}
\usepackage{latexsym}

\usepackage[T1]{fontenc}
\usepackage[utf8]{inputenc}

\usepackage{microtype}

\usepackage{inconsolata}

\usepackage{graphicx}

\usepackage{booktabs}
\usepackage{makecell}
\usepackage{amsmath}
\usepackage{graphicx}
\usepackage{amsfonts}
\usepackage{fancyhdr}
\usepackage{balance}

\newcommand{\eg}{\textit{e.g.}, }
\newcommand{\ie}{\textit{i.e.}, }

\title{YuLan: An Open-source Large Language Model}

\author{%
  Yutao Zhu\thanks{Team leaders.}, Kun Zhou$^{*}$, Kelong Mao, Wentong Chen, Yiding Sun, Zhipeng Chen
  \\ 
  \textbf{Qian Cao, Yihan Wu, Yushuo Chen, Feng Wang, Lei Zhang, Junyi Li, Xiaolei Wang} \\ 
  \textbf{Lei Wang, Beichen Zhang, Zican Dong, Xiaoxue Cheng, Yuhan Chen, Xinyu Tang} \\
  \textbf{Yupeng Hou, Qiangqiang Ren, Xincheng Pang, Shufang Xie} \\
  \textbf{Wayne Xin Zhao, Zhicheng Dou, Jiaxin Mao, Yankai Lin, Ruihua Song, Jun Xu} \\
  \textbf{Xu Chen, Rui Yan, Zhewei Wei, Di Hu, Wenbing Huang, Ze-Feng Gao} \\
  \textbf{Yueguo Chen, Weizheng Lu, \and Ji-Rong Wen} \\
  YuLan team, Renmin University of China \\
  \texttt{\{batmanfly, dou, jrwen\}@ruc.edu.cn}
}

\begin{document}
\maketitle

\thispagestyle{fancy}
\fancyhead{}
\lhead{\raisebox{-0.08cm}{\includegraphics[height=0.6cm]{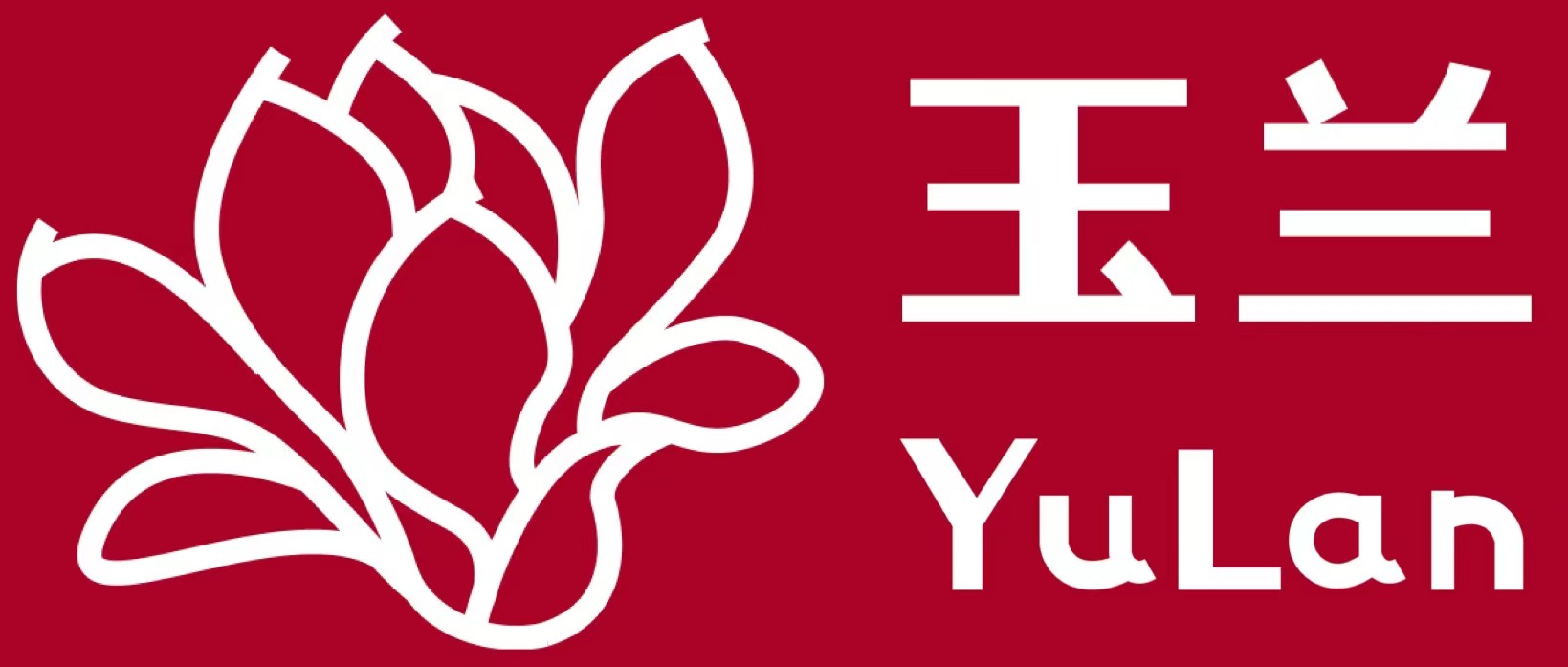}}\vspace{1mm}}

\begin{abstract}
Large language models (LLMs) have become the foundation of many applications, leveraging their extensive capabilities in processing and understanding natural language. While many open-source LLMs have been released with technical reports, the lack of training details hinders further research and development. This paper presents the development of YuLan, a series of open-source LLMs with $12$ billion parameters. The base model of YuLan is pre-trained on approximately $1.7$T tokens derived from a diverse corpus, including massive English, Chinese, and multilingual texts. We design a three-stage pre-training method to enhance YuLan's overall capabilities. Subsequent phases of training incorporate instruction-tuning and human alignment, employing a substantial volume of high-quality synthesized data. To facilitate the learning of complex and long-tail knowledge, we devise a curriculum-learning framework throughout across these stages, which helps LLMs learn knowledge in an easy-to-hard manner. YuLan's training is finished on Jan, 2024 and has achieved performance on par with state-of-the-art LLMs across various English and Chinese benchmarks. This paper outlines a comprehensive technical roadmap for developing LLMs from scratch. Our model and codes are available at \url{https://github.com/RUC-GSAI/YuLan-Chat}.
\end{abstract}

\section{Introduction}
Recent developments in large language models (LLMs) have significantly advanced the field of artificial intelligence~\cite{gpt3,baichuan,palm,llama,llama2,glm-130b,internlm,DBLP:journals/corr/abs-2308-07107}. By scaling up both the size and amount of training data, LLMs have demonstrated emergent capabilities, such as in-context learning~\cite{icl} and chain-of-thought reasoning~\cite{cot}. In-context learning enables LLMs to effectively perform tasks based on a few demonstrations included directly in the prompt without requiring specific model tuning. This capability greatly enhances the practical deployment of LLMs. Furthermore, LLMs' advanced language generation and reasoning capabilities enable them to handle complex tasks across various real-world scenarios, even surpassing human performance in specific tasks~\cite{gpt4}. These advancements have catalyzed a technological revolution in natural language processing (NLP). Typical applications, such as ChatGPT and Copilot, have significantly improved productivity in daily activities.

Most existing LLMs employ decoder-only architectures based on the Transformer~\cite{transformer} model. They are trained in an auto-regressive manner with the objective of next-token prediction. The training process typically includes three stages, namely pre-training, instruction-tuning (also known as supervised fine-tuning), and human alignment. Specifically, during the pre-training stage, LLMs learn natural language and world knowledge from extensive text corpora, laying the foundational understanding of language structure and content. Subsequently, during the instruction-tuning stage, LLMs are trained to interpret and execute human tasks based on natural language instructions. This stage effectively bridges the gap between the objective of pre-training and the specific requirements of practical human tasks. Finally, in the human alignment stage, LLMs are further trained using annotated data that reflects human preferences and values, ensuring the LLMs’ outputs are aligned with human expectations and ethical standards. The training of LLMs is a complex and highly systematic engineering task that involves extensive detail and numerous practical considerations. Despite its importance, there are relatively few references available on this subject. This is primarily due to two factors: First, the research community often lacks the substantial computational resources necessary for training, which limits their ability to thoroughly investigate the training process. Second, the industry often views the details of the training process as proprietary technology and, as such, tends to keep detailed information from the public.

To tackle this problem, we write this report to reveal the detailed training process of our LLM, YuLan. YuLan consists of $12$B parameters and is trained on a vast corpus of English, Chinese, and multilingual data. The model is available in two versions: YuLan-Base and YuLan-Chat. YuLan-Base is trained on approximately $1.7$T tokens text data. Based on this foundation model, YuLan-Chat is further fine-tuned using high-quality synthesized instruction data and is aligned with human preferences through manually annotated data. To improve the overall performance of YuLan, we design several training strategies in different stages. Specifically, (1) during pre-training, we divide the training process into three phases: employing a uniform sampling strategy from a diverse dataset; introducing a capability-enhanced pre-training strategy that adjusts data distribution and expands context lengths to elevate performance on comprehensive benchmarks; and deploying a long-tail knowledge-aware approach to identify and address knowledge gaps in YuLan-Base, significantly boosting its overall capabilities. (2) During the instruction-tuning stage, we also organize the learning process in an easy-to-hard manner. Initially, YuLan is trained with instruction data derived from basic NLP tasks, facilitating its understanding of human-directed tasks. Then, we synthesize more complex instructions and multi-turn dialogue understanding instructions to further improve YuLan's ability to process complex interactions. (3) In the final human alignment stage, we evaluate and select training pairs based on their complexity, adhering to a predefined threshold. This threshold is iteratively adjusted, allowing YuLan-Chat to refine its ability to differentiate and generate high-quality, nuanced text, thereby progressively enhancing its generation quality.

With the aforementioned training strategy, we train YuLan on $96$ NVIDIA A800 GPUs from scratch. The training data include $1.7$T tokens of multilingual texts, $42$M instruction data, and $0.2$M human alignment data. YuLan is evaluated on $22$ public benchmark datasets and achieves comparable performance with several state-of-the-art open-source LLMs. 

\section{Model Architecture}
To be compatible with a variety of toolkits that support the LLaMA~\cite{llama,llama2} model, YuLan follows LLaMA's architecture. Specifically, YuLan has $40$ layers of Transformer decoder with attention heads of $38$. The hidden size is $4,864$, and the feed-forward layer size is $13,056$. In total, YuLan has approximately $12$ billion parameters. 

\begin{table}[t]
    \centering
    \small
    \caption{The compression ratio (Bytes per token) of tokenizers. A higher compression ratio indicates that the text can be tokenized into fewer tokens.}
    \begin{tabular}{lcc}
    \toprule
         & \textbf{LLaMA-2} & \textbf{YuLan} \\
    \midrule
        Chinese (ZH) & 1.94 & 2.15 \\
        English (EN) & 4.10 & 4.10 \\
        Code & 2.79 & 2.80 \\
        Academic papers & 3.88 & 3.88 \\
        Average (EN \& ZH) & 3.02 & 3.13 \\
    \bottomrule
    \end{tabular}
    \label{tab:compress}
\end{table}

\paragraph{Tokenizer}
Optimizing the tokenizer is a crucial aspect of model training, particularly when considering the efficiency of text compression and inference. A larger vocabulary generally enables better text compression rates but requires more data and resources for effective training. The original LLaMA tokenizer, while effective for English texts, exhibit limitations when processing Chinese texts, often breaking down individual Chinese characters into multiple tokens. This inefficiency complicates the encoding process and restricts the maximum input length for Chinese texts. To address these challenges, we enhance the original LLaMA tokenizer by expanding its vocabulary to include additional tokens specifically for Chinese. This expansion preserves the original English tokens and integrates new tokens derived from the WordPiece algorithm applied to a Chinese text subset from our pre-training data. This method ensures the tokenizer's improved performance on Chinese texts without degrading its effectiveness on non-Chinese texts. As a result, the updated tokenizer contains a total of $51,190$ tokens, and we pad it to $51,200$ to enhance training efficiency. The compression ratios are shown in Table~\ref{tab:compress}.

\paragraph{Positional Embeddings}
Following LLaMA, we use rotary position embeddings (RoPE)~\cite{rope}. RoPE utilizes a rotation matrix to encode absolute positions while incorporating relative positional dependencies within the self-attention mechanism. This design not only allows for variable sequence lengths but also introduces a decay effect in the inter-token dependencies as the relative distances increase. Besides, RoPE benefits from compatibility with Flash Attention, which significantly enhances training speed.

\paragraph{Activation and Normalization}
We use SwiGLU~\cite{swiglu} as the activation function. It is a variant of gated linear units that incorporates Swish functions~\cite{swish} as the non-linear activation. Compared to the vanilla MLPs that use two matrix multiplications, SwiGLU uses three. Therefore, to keep the number of parameters and the amount of the computation the same, the MLP layer's hidden size is reduced to $\frac{2}{3}4d$. This adjustment ensures that the parameter count in SwiGLU's three matrices remains comparable to that in the traditional two-matrix configuration of vanilla MLPs. As for the layer normalization, we apply RMSNorm~\cite{rmsnorm} to the input of each Transformer sub-layer to improve the training stability. The normalization hyper-parameter epsilon is set as $1.0e$-$6$.

\paragraph{Maximum Input Length}
With the development of LLMs, the maximum input length is also increasing. Unfortunately, since our computational resources are very limited, we cannot training LLMs with long context from scratch. Therefore, we follow the idea of XGen~\cite{xgen} and train YuLan with increasing maximum length. Initially, YuLan is trained on sequences up to $2,048$ tokens for the first $600$B tokens, subsequently increasing to $4,096$ tokens. By this means, we save a lot of training time and achieve high performance. 

\paragraph{Optimization}
We use GPT-NeoX framework~\cite{gpt-neox-library} for training, which integrates Megatron-LM~\cite{arxiv19_megtronlm} and DeepSpeed.\footnote{\url{https://github.com/microsoft/DeepSpeed}}
YuLan is trained with AdamW optimizer~\cite{adamw}, with the hyper-parameters $\beta_1=0.9$ and $\beta_2=0.95$. A cosine learning rate schedule is applied, and the final learning rate is $10\%$ of the maximum learning rate ($3e$-$4$). We implement a weight decay of $0.1$ and gradient clipping at $1.0$, with an initial warm-up phase comprising $0.1\%$ of total training steps. YuLan is trained using BFloat16 mixed precision to enhance handling of large numerical values, which is critical for LLM training stability. The checkpoint activation technique is applied to save memory. To train our model with a proper batch size ($4$M tokens), we apply both ZeRO-powered data parallelism and tensor parallelism. We find that the ZeRO stage one with tensor parallelism as two is optimal for our training. Flash attention is also applied to accelerate training. By integrating all these strategies, our training achieves around $180$ TFLOPS on $96$ NVIDIA A800 GPUs. 

\section{Pre-training}
We pre-train YuLan-Base on a mixture of Chinese, English, and multi-lingual data from diverse domains. For multilingual data, we follow PaLM~\cite{palm}, LLaMA~\cite{llama}, and CC-100~\cite{cc100} and then select the following languages: \texttt{de}, \texttt{fr}, \texttt{es}, \texttt{pl}, \texttt{it}, \texttt{nl}, \texttt{tr}, \texttt{pt}, \texttt{ru}, \texttt{fi}, \texttt{cs}, \texttt{ja}, \texttt{no}, \texttt{ko}, \texttt{da}, \texttt{id}, \texttt{ar}, \texttt{uk}, \texttt{ca}, \texttt{hu}, \texttt{ro}, \texttt{fa}, \texttt{bg}, \texttt{el}, \texttt{he}, \texttt{hi}, \texttt{hr}. In this section, we will first introduce the dataset we collect and pre-process for pre-training, and then introduce our pre-training strategies.

\begin{table*}[t]
    \centering
    \small
    \caption{Overview of pre-training datasets. Raw size and weighted size are the numbers of tokens before and after sampling, respectively. \# Epoch is the number of passes over each constituent dataset during a full epoch over the final dataset. Weight is the percentage of bytes in the final dataset occupied by each dataset.}
    \begin{tabular}{lrrrrrrr}
    \toprule
        \textbf{Dataset} & \textbf{English} & \textbf{Chinese} & \textbf{Multilingual} & \textbf{Raw Size} & \textbf{\# Epoch} & \textbf{Weighted Size} & \textbf{Weight} \\
    \midrule
        Web pages & $\checkmark$ & $\checkmark$ & $\checkmark$ & $1,220$B & $1$ & $1,220$B & $72.6\%$ \\
        Code & $\checkmark$ & & & $101$B & $1$ & $101$B & $6.0\%$ \\
        Encyclopedia & $\checkmark$ & $\checkmark$ & $\checkmark$ & $18$B & $3$ & $54$B & 3.2\% \\
        % Wikipedia & $\checkmark$ & $\checkmark$ & $\checkmark$ & 12B & 3 & 2.1\% \\
        Academic papers & $\checkmark$ & & & $50$B & $1$ & $50$B & $3.0\%$ \\
        QA Forums & $\checkmark$ & & & $26$B & $1$ & $26$B & $1.5\%$ \\
        % Stack Exchange & $\checkmark$ & & & 5B & 1 & 0.3\%\\
        Books & $\checkmark$ & $\checkmark$ & & $43.75$B & $2$ & $87.5$B & $5.3\%$ \\
        News articles & $\checkmark$ & $\checkmark$ & $\checkmark$ & $134$B & $1$ & $134$B & $8.0\%$ \\
        % Chinese QA & & $\checkmark$ & & 21B & 1 & 1.2\%\\
        Legal documents & & $\checkmark$ & & $3$B & $1$ & $3$B & $0.2\%$ \\
        % Encyclopedia & & $\checkmark$ & & 6B & 3 & 1.1\% \\
        Patents & & $\checkmark$ & & $2$B & $1$ & $2$B & $0.1\%$ \\
        Educational assessments & & $\checkmark$ & & $1.25$B & $2$ & $2.5$B & $0.1\%$ \\
    \midrule 
        Total & - & - & - & - & - & $1,680$B  & $100\%$ \\
    \bottomrule
    \end{tabular}
    % }
    \label{tab:data_stat}
\end{table*}

\subsection{Pre-training Data}
% In general, we categorize our data into web pages, code, encyclopedia, research paper, question-answering (QA) forums, books, news, legal, patent, and exam. The statistics of the data are provided in Table~\ref{tab:data_stat}. Below, we provide the collecting and pre-processing details to support future research. 
For the development of YuLan-Base, we systematically organize the pre-training data into distinct categories: web pages, code, encyclopedia, academic papers, question-answering (QA) forums, books, news articles, legal documents, patents, and educational assessments. Detailed statistics regarding the volume and language of these data sources are presented in Table~\ref{tab:data_stat}. In the following sections, we introduce the data collection we used. To facilitate reproducibility and further research in this field, we provide the details of data pre-processing in Appendix~\ref{app:data}. Our pre-processing tool YuLan-GARDEN~\cite{yulan-garden} has been released.\footnote{\url{https://github.com/RUC-GSAI/Yulan-GARDEN}}

\paragraph{Web Pages}
Web pages offer a broad spectrum of knowledge across various domains, making them essential for developing models that are robust and capable of understanding context in multiple fields. Specifically, our dataset includes data from OpenWebText2~\cite{thepile}, C4~\cite{DBLP:conf/emnlp/DodgeSMAIGM021}, RefinedWeb~\cite{refinedweb}, CC-100~\cite{cc100}, ClueWeb 22~\cite{clueweb22}, CC-Stories~\cite{ccstories}, and Dolma's CC~\cite{dolma}. In addition to these sources, we process raw data from Common Crawl (CC) dumps, particularly focusing on events that occurred between January 2021 and February 2023. To manage the practical challenges of HTML content extraction and constraints of disk space, we utilize the WET file format, which includes only plain text, for further preprocessing. We selectively retain texts in English, Chinese, and other multilingual texts that are specified in our language list, ensuring a diverse yet controlled dataset for model training.

\paragraph{Code}
Incorporating programming code into pre-training data is critical for enhancing the capabilities of LLMs, particularly in fostering the development of an emergent chain-of-thought and algorithmic reasoning. Code inherently embodies structured, logical thinking and provides a sequential understanding of tasks, which are fundamental to developing LLMs that can emulate human-like problem-solving skills. Studies have shown that the inclusion of programming code not only augments the syntactic understanding but also significantly boosts the model's ability to perform complex reasoning and execute task-specific functions~\cite{gpt3, zhao2023survey}. Hence, our dataset extensively incorporates code from various sources to cultivate these advanced capabilities in our LLM. We source the programming code from two primary repositories: the Stack~\cite{stack} and GitHub.

\paragraph{Encyclopedia}
Encyclopedias represent a cornerstone resource in the pre-training of LLMs, offering a vast repository of structured, high-quality human knowledge essential for building comprehensive understanding. These resources are pivotal in enhancing the factual accuracy and depth of knowledge of LLMs, making them necessary for applications requiring reliable information and nuanced content generation. In our pre-training, we extend beyond the conventional use of Wikipedia to include the Baidu Encyclopedia, thereby enriching our dataset with expansive Chinese linguistic and cultural knowledge.

\paragraph{Academic Papers}
Academic papers are a pivotal source for the pre-training of LLMs due to their complex structure, formal language, and rich scientific content. These documents provide a diverse array of knowledge and are instrumental in enhancing the reasoning capabilities of LLMs, allowing them to perform more effectively in tasks requiring deep understanding and analytical skills. To this end, we incorporate a substantial corpus of papers from two major repositories: arXiv and the peS2o dataset~\cite{peS2o}.

\paragraph{QA Forums}
Question-answering datasets are crucial for the pre-training of LLMs, as they provide the necessary supervisory signals for LLMs and promote the improvement of the models' capabilities in language understanding, knowledge acquisition, context awareness, generalization, and dialogue generation. The improvement of these capabilities is crucial for developing more intelligent, efficient, and practical LLMs. We use the Stack Exchange (in English) dataset and the Zhihu (in Chinese) dataset.

\paragraph{Books}
Books represent an invaluable data source for training LLMs, especially in fostering an understanding of long context dependency in natural language processing. High-quality books provide structured and detailed content that is crucial for enhancing the depth and scope of the model's comprehension capabilities. Particularly, textbooks have been proven to be exceptionally effective in improving LLMs' performance due to their rich, authoritative, and well-organized content~\cite{DBLP:journals/corr/abs-2306-11644,DBLP:journals/corr/abs-2309-05463}. Our pre-training dataset includes a diverse selection from the Books3 dataset, Project Gutenberg, CBook, Bestsellers, English textbooks, and Chinese textbooks, each offering unique advantages to the training process.

\paragraph{News Articles}
News provides a stream of current events and real-time data that is crucial for training LLMs to be relevant and responsive to the latest global developments. By integrating news data from diverse sources, LLMs can better grasp the nuances of journalistic language, adapt to varying narrative styles, and improve their accuracy in information retrieval and generation tasks. In our dataset compilation, we include news from CC-news, RealNews~\cite{realnews}, and the news articles from China International Communications Group (CICG) to cover a wide range of topics and perspectives.

\paragraph{Legal Documents}
Legal judgment documents are also helpful for training LLMs due to their formal, structured nature and the logical complexity they embody. These documents encapsulate rigorous reasoning processes and legal terminology, making them beneficial for enhancing the analytical capabilities of LLMs. The precision and clarity required in legal language training help improve the model's ability to understand and generate text within specific, rule-based contexts, which is pivotal for applications in legal assistance, automated compliance checks, and advanced query-response systems in the legal domain.

\paragraph{Patents}
Patent applications are also useful for training LLMs due to their standardized format and formal, technical language. These documents are rich in specialized vocabulary and complex sentence structures, reflecting high levels of precision and clarity. Training LLMs on such data can significantly enhance their ability to parse and generate text within technical contexts. 

\paragraph{Educational Assessments}
Educational assessments provide structured problem-solving environments that are vastly different from general text data, helping models learn to navigate and understand the specific formats and logical reasoning required in standardized testing. The inclusion of this type of data trains LLMs not only in content knowledge but also in the application of this knowledge within the constraints of a given question structure, which is crucial for achieving high performance on standardized assessments like MMLU~\cite{hendryckstest2021}, C-Eval~\cite{huang2023ceval}, and AGIEval~\cite{zhong2023agieval}.

% \subsubsection{Chinese Encyclopedia}
% % Yiding Sun

% \subsubsection{Chinese Web Pages}
% % Qian Cao

% Chinese web page data is a majority of the source of Chinese corpus, and it is mainly contained in Common Crawl.
% Thus, we took the RefinedWeb, CC-100, ClueWeb22, and our own constructed CC dataset to compose this.
% Preprocessing details follow those in ~\ref{sec.data.cc}.

% \subsubsection{Chinese Books}
% % Lei Zhang, Yihan Wu, Feng Wang, Yiding Sun

% \subsubsection{Chinese News}
% % Yiding Sun, Feng Wang

% \paragraph{CICG-zh}
    
% Finally, we unified paragraph breaks into two consecutive `\textbackslash n', and keep newlines as a single `\textbackslash n'.

% \subsubsection{Multilingual Data}

% \subsubsection{Multilingual Wikipedia}
% Wentong Chen

% \subsubsection{Multilingual Web Pages}
% % Qian Cao

% Like the web pages in other languages, the multilingual web page data contains several sources including the RefinedWeb, CC-100, ClueWeb22, and our own constructed CC dataset, with the processing details presented in~\ref{sec.data.cc}.
% We retain the intersection of the top 30 languages from PaLM~\cite{palm}, LLaMA~\cite{llama}, and CC-100~\cite{cc100} respectively, which can be processed by CCNet.

% \subsubsection{Multilingual News}
% % Yihan Wu
% \paragraph{CICG-ml} 
% China International Communications Group~(CICG) is an official foreign-language publishing and communication organization which provides reliable multilingual news. Similar to multilingual web pages, we retain Top 30 languages of multilingual news.

\begin{table*}[t]
    \small
    \centering
    \caption{Training settings for pre-training in different stages. EN: English, ZH: Chinese, ML: Multilingual.}
    \begin{tabular}{ccccccc}
        \toprule
        \textbf{Stage}   &  \makecell[c]{\textbf{Data Distribution}\\ \textbf{(EN:ZH:ML)}} & \textbf{Context length} & \textbf{\# Tokens} & \textbf{Initial LR} & \textbf{Min LR} & \textbf{Batch size} \\ 
        \midrule
        Stage 1 & $76:22:2$ & $2,048$ & $600$B & $3e$-$4$ & $3e$-$5$ & $4$M    \\
        Stage 2 & $90:10:0$ & $4,096$ & $900$B & $2e$-$5$ & $2e$-$5$ & $4$M    \\
        Stage 3 & $62:33:5$ & $4,096$ & $180$B & $2e$-$5$ & $2e$-$5$ & $4$M    \\ 
        \bottomrule
    \end{tabular}
    \label{tab:pre-training_data}
\end{table*}

\subsection{Pre-training Process}
\label{sec:curriculum_pre-training}
Our pre-training process can be divided into three stages: (1) standard pre-training; (2) capability-enhanced pre-training; and (3) long-tail knowledge-aware pre-training. In the first stage, we follow existing studies and apply a standard training strategy, which involves predicting the next token on randomly sampled data. Then, we notice a plateau in performance improvement and intermittent instability, so we refine our approach to enhance YuLan-Base's overall capability. Finally, we design a approach to detect and augment the YuLan-Base's comprehension of long-tail knowledge, thereby reducing inaccuracies and improving task-specific performance in downstream applications. The training settings of these stages are provided in Table~\ref{tab:pre-training_data}.
% We propose a curriculum-based pre-training approach aimed at iteratively enhancing the model's comprehension of long-tail knowledge. This method involves identifying areas where the model lacks proficiency in long-tail knowledge and employing the TF-IDF algorithm to retrieve relevant pre-training data. By iteratively incorporating this data into the training curriculum, we aim to reinforce the model's understanding of long-tail knowledge and mitigate hallucinations in downstream tasks.

\begin{figure}
    \centering
    \includegraphics[width=\linewidth]{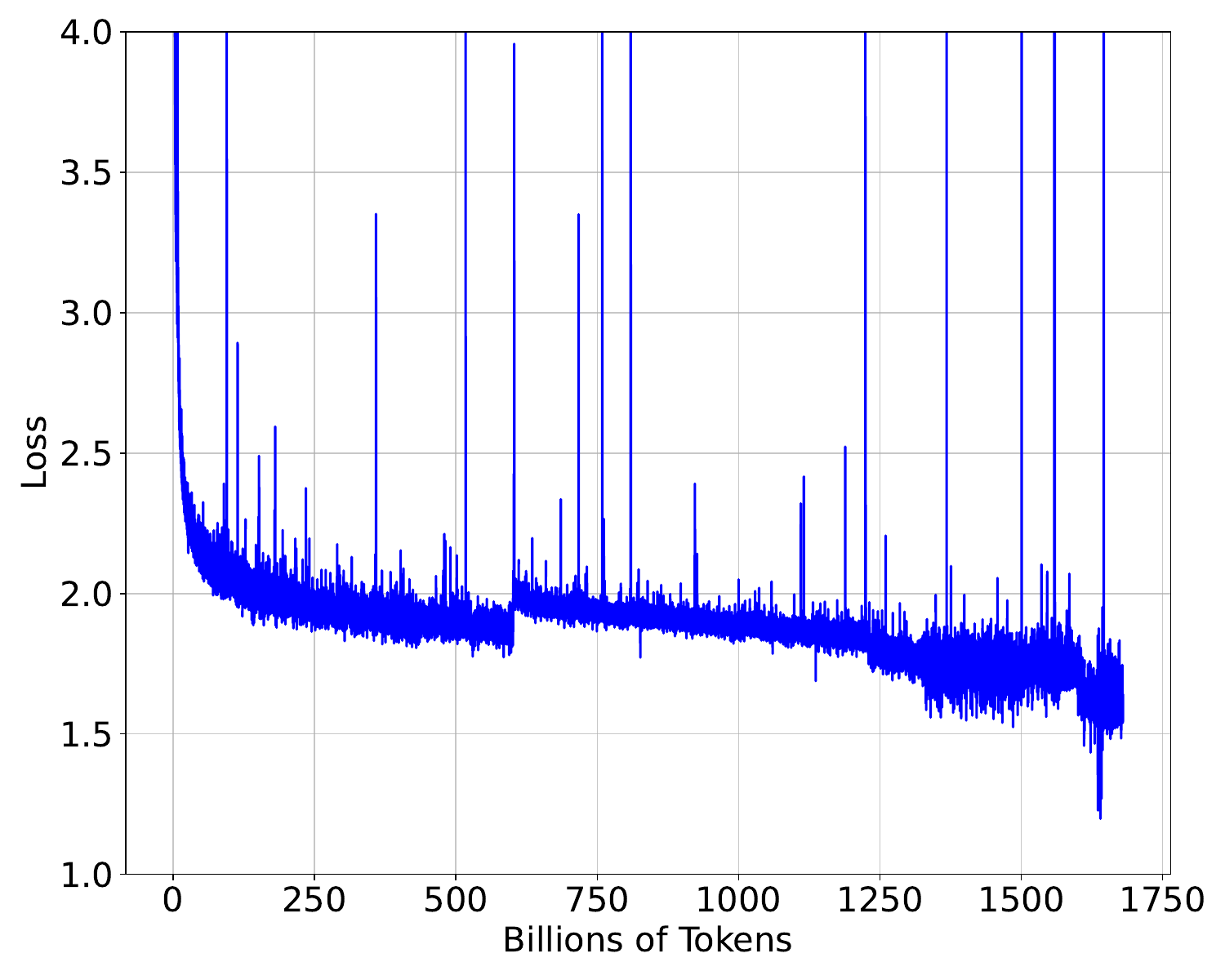}
    \caption{The pre-training loss of YuLan-Base.}
    \label{fig:loss}
\end{figure}

\subsubsection{Standard Pre-training}
In standard pre-training, we train YuLan-Base with the next-token prediction objective, which is defined as:
\begin{align}
    p_{\text{LM}} = \prod_{i=1}^{n}p_{\theta}(x_i|x_{<i}),
\end{align}
where $x_{<i}$ denotes the sequence of tokens preceding $x_i$ at each step, and $\theta$ represents the parameters of the model. We mix and randomly sample all training data to construct data batches at each step. The maximum context length is set as $2,048$ tokens at this stage. Note that we follow the training strategy of GPT-2~\cite{gpt2}, where data from the same source are concatenated into long sequences and segmented into training samples of equal length ($2,048$ tokens). This method avoids the need for zero-padding within text sequences, thereby enhancing training efficiency. Throughout this stage, we observe a consistent decrease in training loss and a gradual improvement in model performance.

\subsubsection{Capability-Enhanced Pre-training}
Following the standard pre-training with $600$B tokens, we observe fluctuations in the model's performance on certain benchmarks, notably on comprehensive benchmarks such as the MMLU, where results approach those of random chance. To tackle this problem, we conduct a series of empirical studies (detailed in Section~\ref{sec:exam}) that reveal the significant impact of incorporating educational assessments into the pre-training process. This enhancement is particularly beneficial for performance on comprehensive benchmarks for several reasons: (1) Benchmarks like MMLU consist of multiple-choice questions, a format seldom appeared in natural language texts. Educational assessments frequently contain such questions, aiding the model in task familiarization. (2) These benchmarks often resemble closed-book quizzes that challenge the model to respond based solely on its acquired knowledge. Educational assessments not only present the correct answers but also elaborate on the reasoning and analysis behind them, thereby effectively guiding the model in applying its inherent knowledge. Furthermore, we increase the maximum context length to $4,096$ tokens, enhancing the YuLan-Base's performance in comprehending long documents. These strategic modifications lead to consistent performance improvements across all evaluated benchmarks.

\subsubsection{Long-tail Knowledge-Aware Pre-training}
After pre-training on $1,500$B tokens, YuLan-Base achieves performance on par with many popular open-source LLMs across various benchmarks. However, post-instruction tuning evaluation reveals deficiencies in handling certain long-tail knowledge topics. To mitigate these problems, we propose a strategy to identify areas of knowledge that the model has not effectively learned. Our approach involves augmenting our pre-training dataset with additional relevant content specifically targeted at these identified gaps, thereby improving the model's ability to process and understand long-tail knowledge.

\paragraph{Weak Long-tail Knowledge Detection}
The first step is to identify which knowledge is incompletely understand by YuLan-Base. Given the complex nature of knowledge, we focus on \textit{entities} as a proxy for evaluating the model's understanding of relevant knowledge. Inspired by recent studies~\cite{self-ask}, we propose synthesizing question-answer pairs that evaluate the model's retention of entity-specific knowledge. These questions are crafted to test the model's comprehension at the entity level, utilizing entities and their descriptions from encyclopedic sources such as Wikipedia. The rationale is straightforward: if the model fails to accurately respond to questions about a particular entity,  it indicates a gap in the acquisition of relevant knowledge. Due to YuLan-Base's difficulties in following human instructions and producing effective responses, we enhance its performance by fine-tuning it with a selected subset of our instruction tuning dataset, resulting in a temporarily improved version, YuLan-tmp. We then employ YuLan-tmp to identify deficiencies in the model's understanding of less commonly addressed, long-tail knowledge.

Specifically, we first construct an entity list from our encyclopedia datasets, including Wikipedia and Baidu Encyclopedia. To ensure data quality, we exclude entities that are either briefly described or infrequently mentioned within these datasets. Then, we manually craft several templates and employ other advanced LLMs (\ie{} ChatGPT) to generate questions related to these entities. For each entity $v$ along and its detailed descriptions $d_v$ in the encyclopedia, we perform string matching to identify entities frequently co-occurring with $v$ in $d_v$. This process helps us establish a related entity set $V$. Using $d_v$ and the relationships identified within $V$,  we then generate questions $q$ about $v$ or its interactions with other entities in $V$. For example, for the entity ``Emperor Taizong of Tang'', potential questions might include, ``Could you provide some context about Zhenguan's Enlightened Administration?'' or ``Could you elaborate on the connection between Emperor Taizong of Tang and Empress Wu Zetian?'' After formulating a substantial number of questions, we pair each with its corresponding article $d_v$ and submit them to ChatGPT to generate reference answers $a$. These question-answer pairs $\{(q_i, a_i)\}_{i=1}^{M_v}$ are accumulated for each entity, where $M_v$ denotes the total number of pairs for the entity $v$.

In the knowledge detection process, each question $q_i$ is input to YuLan-tmp to generate a response $a_i'$. Both the generated response and the reference answer are then fed into another LLMs for evaluation, which provides binary feedback on their alignment.\footnote{This task is relatively simple, so we use Baichuan-2-13B instead of ChatGPT for saving costs.}  This feedback facilitates the calculation of YuLan-Base's understanding of each entity using the following scoring metric:
\begin{align}
    s_v = \frac{1}{M_v} \sum_{i=1}^{M_v} f(q_i, a_i, a_i').
\end{align}
By setting an appropriate threshold $\epsilon$, we identify the set of entities $V' = \{v | s_v < \epsilon, v \in V\}$ that are not adequately understood by the current YuLan-Base model.

\paragraph{Relevant Knowledge Retrieval}
Given the identified set of entities $V'$ that the YuLan-Base inadequately understands, we extract relevant data from the pre-training dataset $D_{\text{pre}}$ to mitigate these gaps. Considering the large scale of the entire $D_{\text{pre}}$, we employ the efficient TF-IDF algorithm to measure the similarity between each sample in $D_{\text{pre}}$ and the entity-related questions. We then retrieve the top-$k$ samples that have the highest similarity for each entity. The duplicated samples are removed to obtain the dataset $D_{\text{pre}}'$. This refined subset, $D_{\text{pre}}'$, is tailored specifically to enhance the model's comprehension of the entities it previously struggled with.

\paragraph{Multi-round Iterative Training}
After pre-training on $D_{\text{pre}}'$, YuLan-Base's performance on long-tail knowledge can be improved. This process can be repeated multiple times to iteratively enhance the model. Specifically, in each iteration, a new entity set $V$ along with associated question-answer pairs are synthesized to identify gaps in the model's current knowledge. This process led to the identification of a refined set of entities $V'$, which the model struggles with. A new dataset $D_{pre}'$ is then constructed using TF-IDF based on the questions of $V'$. Finally, the model undergo further pre-training on this updated dataset. This iterative pre-training cycle is repeated five times, and we cannot observe significant performance improvement on these entity-related questions. 

During this stage of pre-training, all Chinese data and about a half of English data are selected based on our designed strategy to improve YuLan-Base's performance on areas of weak long-tail knowledge. The remaining portion of data is still randomly sampled from our pre-training dataset. This pre-training strategy is designed to enhance the model's capability in handling user input involving less frequent knowledge.

% \subsection{Training Details}
% We adopt the Megatron-LM framework~\cite{arxiv19_megtronlm} with DeepSpeed~\cite{kdd20_deepspeed} for highly efficient pre-training. The pre-training was conducted on 96 NVIDIA A800 80G GPUs across 12 nodes. The model is trained with the standard language model loss on a mixture of 1680 billion tokens from the specified data sources. We ensure that the training order of samples from the same data source matches their order in the original text, allowing the model to learn continuous knowledge between adjacent batches as much as possible.

% We use the AdamW optimizer~\cite{iclr19_adamw} with cosine learning rate decay. Our pre-training is split into three stages, each with different learning rates, data distributions, and context window sizes. The training settings for the different stages are summarized in Table~\ref{tab:pre-training_data}. In particular, we implement curriculum learning to improve the model's memory of long-tail knowledge and reduce hallucinations in downstream tasks in the third stage (i.e., the last stage). We provide more details on this in Section~\ref{sec:curriculum_pre-training}.

\section{Supervised Fine-tuning and Human Alignment}
During pre-training, we focus on training YuLan to accurately predict the next token in text sequences. Following this, we implement a supervised fine-tuning process (also known as instruction tuning), which adapts the model to understand and execute human-like tasks. To optimize this learning, we employ a curriculum-based approach that systematically organizes instruction data from simpler to more complex tasks. Following fine-tuning, we perform human alignment learning to ensure the model's outputs align with human values. This includes adjusting the training methodology by controlling the similarity between positive and negative samples. Such control allows the model to progressively learn to discern finer distinctions between samples, ultimately achieving better alignment with human preferences.

\subsection{Curriculum Instruction-Tuning}
The target of instruction-tuning is to transfer the learning objective of LLMs from predicting next token to tackling real human tasks. While many instruction datasets have been released, they mainly focus on single-turn tasks or simple multi-turn tasks, which limit the model's ability to learn complex, context-dependent tasks. To tackle this challenge, we first collect existing instruction dataset, and then we synthesize more multi-turn instructions based on existing data. Finally, we design a curriculum to fine-tune YuLan to learn from simple to more complex instructions. This curriculum learning-based training process allows for incremental improvements in YuLan's performance, enhancing its capability to handle intricate tasks that require advanced contextual reasoning.

% After pre-training, we perform instruction-tuning on YuLan.
% Considering the complexity of real-world instructions, we mainly focus on improving the capability of YuLan on understanding and following complex instructions.
% Therefore, we need to enrich the dataset by crafting new complex instructions, and then employ a curriculum learning approach to training YuLan from simple to complex ones, \eg single-turn simple ones to multi-turn complex reasoning instructions.
% Concretely, we first synthesize complex multi-turn instructions using ChatGPT, then devise a complex estimation function to divide the synthetic and collected ones into simple or difficult categories. 
% Finally, we train YuLan on simpler instructions, and then transitions to complex ones, mimicking the learning process of humans in real world.

\paragraph{Instruction Data Collection}
We first collect instruction datasets that have been widely used for instruction tuning. These data cover various natural language tasks or tasks in real-world applications. We consider two primary categories for data collection: (1) To improve YuLan's fundamental capabilities, such as knowledge utilization and reasoning, we use datasets including Flan-v2~\cite{longpre2023flan}, OpenOrca~\cite{OpenOrca,mukherjee2023orca}, Chinese data in xP3~\cite{muennighoff2022crosslingual}, MetaMathQA~\cite{yu2023metamath}, and  MathInstruct~\cite{yue2023mammoth}. Additionally, to enhance YuLan's comprehension of Chinese factual knowledge, we synthesize instructions via ChatGPT based on entities from Baidu Encyclopedia. (2) To improve YuLan's ability to follow instructions, we incorporate the  ShareGPT~\cite{vicuna2023}, which contains multi-turn instructions.

% construct a complex instruction dataset to improve the basic abilities of YuLan , and a multi-turn instruction dataset to improve its instruction-following ability. 
% We adopt Flan v2~\cite{longpre2023flan}, OpenOrca~\cite{OpenOrca,mukherjee2023orca}, Chinese data in xP3~\cite{muennighoff2022crosslingual}, and the examination problems of High school in China as the basic components of the complex instruction dataset.
% 
% Moreover, we adopt MetaMathQA~\cite{yu2023metamath} and MathInstruct~\cite{yue2023mammoth} as the complex instructions to improve the complex reasoning ability of YuLan.
% These problems are added to the complex instruction dataset.
% For the multi-turn instruction dataset, we utilize ShareGPT~\cite{vicuna2023} and collect the synthetic data from ChatGPT and GPT-4.

\paragraph{Complex Multi-turn Instruction Synthesis}
In addition to collect instruction data from existing datasets, we also synthesize some complex multi-turn instructions. However, directly synthesize complex multi-turn instructions is very challenging. Therefore, we adopt a multi-stage approach to increase complexity based on existing instruction data. The synthesis process involves three stages: instruction merging, multi-turn conversion, and complexity enhancement. 

(1) \textit{Instruction merging.} We begin by collecting instruction datasets from the open-source community (\ie WizardLM-Instruct and Alpaca), removing duplicates to form a base set of instructions.\footnote{\url{https://github.com/nlpxucan/WizardLM}, \url{https://github.com/tatsu-lab/stanford_alpaca}}
Employing the TF-IDF algorithm, we determine the similarity between instructions and select pairs with high similarity. These pairs are then merged using a prompt to ChatGPT: ``\emph{Please merge the following two semantically similar instructions into a new instruction that incorporates the functionalities of both instructions and is more complex.}'' This ensures that the merged instructions retain semantic similarity and increased complexity. The merged instruction is subsequently input into ChatGPT to generate an appropriate response.

(2) \textit{Multi-turn conversion.} The next phase involves converting the merged instructions into multi-turn instructions to further enhance their complexity. To ensure the diversity of topics, we collect a set of 293 topics from chat communities (\ie Zhihu and Reddit). For each merged instruction, we utilize ChatGPT again to generate a next-turn question in terms of a randomly selected topic. An example prompt is: ``\emph{Please generate a question related to the topic `modern history' and ensure its consistency with the context of the conversation.}'' 

(3) \textit{Complexity enhancement.}
To ensure the generated instructions are sufficiently complex, we use a prompt that encourages ChatGPT for deeper and broader knowledge exploration: ``\emph{Please modify the following question into a more complex instruction that significantly enhances the depth and width of the involved knowledge.}'' This process yields highly complex instructions which are then processed through ChatGPT to generate responses. Through the above process, we can obtain the complex instructions, which are fed into ChatGPT to generate responses. Following a quality assessment, the refined set of synthetic complex multi-turn instructions is compiled into a dataset.

\paragraph{Simple-to-Complex Curriculum}
Based on the collected open-source instruction datasets and our synthesized complex instruction dataset (around $41$M instructions in total), we combine and re-split them into two parts based on their complexity: a simple set and a complex set.\footnote{Except for ShareGPT, which contains multi-turn instructions, we directly categorize it into the complex set.} The complexity of each instruction is measured using the following equation:
\begin{align}
    \text{Comp}(x,y) =& \lambda_1 \cdot L_{\text{turn}} + \lambda_2 \cdot L_{\text{length}} \notag \\
    &+ \lambda_3 \cdot \text{Loss}_{\text{it}}(x, y). \label{eq-comp}
\end{align}
Here, $\lambda_1$, $\lambda_2$, and $\lambda_3$ are the hyperparameters; $L_{\text{turn}}$ and $L_{\text{length}}$ denote the number of turns and the length of the instruction; $\text{Loss}_{\text{it}}(x, y)$ is the loss calculated by the current model:
\begin{equation}
\text{Loss}_{\text{it}}(x, y) = \sum_{i=1}^{|y|} \log P(y_i | x, y_{1:i-1}),    \label{eq-sft-loss}
\end{equation}
where $y_i$ represents the $i$-th token in the output $y$, and $y_{1:i-1}$ denotes the sequence up to the $i-1$ tokens. 
Based on the complexity value computed by Equation~(\ref{eq-comp}), we set a threshold to categorize all instruction data into either the simple or complex set.
The training starts from the simple instruction set and progresses to the complex set. This structured curriculum allows YuLan-Chat to incrementally acquire and apply knowledge from the instructions, enhancing its capability to comprehend and execute more complex instructions efficiently.
% Thereafter, we first train YuLan on the simple instruction set, and then on the difficult set.
% Such a simple-to-complex curriculum enables YuLan to smoothly learn all the knowledge from the instructions, help it better understand and follow complex instructions.
% We also utilize the loss function as Eq.~\ref{eq-sft-loss} to train YuLan for predicting each token in the output $y$ based on the input instruction $x$.

\subsection{Curriculum Human Alignment Learning}
After instruction tuning, we further enhance our YuLan for better human alignment. This stages focuses on strengthening its capability of distinguishing subtle negative inputs (\eg obscure abuse), and avoiding generating outputs conflicting with human values. Despite the abundance of open-source human alignment datasets, the complexity of instances within these datasets varies considerably. To address this, we implement a reward function based on direct preference optimization (DPO) to measure instance difficulty and design an easy-to-hard curriculum for model training.

\begin{table*}[t]
    \centering
    \small
    \caption{Overview of datasets and benchmarks for evaluation. The designated datasets, delineated by underlines, constitute the data utilized for our evaluation purposes. Primarily, we employ the test set for conducting our evaluation. In instances where answers are unavailable within the test set, we resort to utilizing the validation set.}
    \setlength\tabcolsep{0.9mm}{
    \begin{tabular}{llcrrr}
    \toprule
        \textbf{Dataset} & \textbf{Type} & \textbf{Language} & \textbf{Train Set} & \textbf{Vaild. Set} & \textbf{Test Set}\\
    \midrule
        BoolQ~\cite{clark2019boolq} & Natural Language Inference &EN &9,427 &\underline{3,270} &3,245 \\
        PIQA~\cite{Bisk2020PIQA} &Physical Commonsense Reasoning &EN &16,113 &\underline{1,838} &3,084 \\
        Hellaswag~\cite{zellers2019hellaswag} &Commonsense NLI &EN &39,905 &\underline{10,042} &10,003 \\
        WinoGrande~\cite{ai2:winogrande} &Winograd Schema Challenge &EN &40,398 &\underline{1,267} &1,767 \\
        WSC273~\cite{levesque2012winograd} &Winograd Schema Challenge &EN &- &- &\underline{273} \\
        ARC-easy~\cite{allenai:arc} &Science Questions &EN &2,251 &570 &\underline{2,376} \\
        ARC-challenge~\cite{allenai:arc} &Science Questions &EN &1,119 &299 &\underline{1,172} \\
        OpenBookQA~\cite{OpenBookQA2018} &Common Sense Knowledge &EN &4,957 &500 &\underline{500} \\
        CommonSenseQA~\cite{talmor-etal-2019-commonsenseqa} &Common Sense Reasoning &EN &9,741 &1,221 &\underline{1,140} \\
    \midrule
        TriviaQA~\cite{2017arXivtriviaqa} &Factual Knowledge &EN & 138,384& 17,944 & \underline{17,210}\\
    \midrule
        CoQA~\cite{reddy-etal-2019-coqa} &Conversational Question Answering &EN &7,199 &\underline{500} &- \\
        RACE-middle~\cite{lai-etal-2017-race} &Chinese Middle School English Exams &EN &25,421 &1,436 &\underline{1,436} \\
        RACE-high~\cite{lai-etal-2017-race} &Chinese High School English Exams &EN &62,445 &3,451 &\underline{3,498} \\
        CMRC2018~\cite{cui-emnlp2019-cmrc2018} &Span-Extraction Chinese MRC &ZH &10,142 &3,219 &\underline{1,002} \\
        C3-Dialogue~\cite{sun2019investigating} &Multiple-Choice (Dialogues) &ZH &4,885 &1,628 &\underline{1,627} \\
        C3-Mix~\cite{sun2019investigating} &Multiple-Choice (Mixed-Genre Texts) &ZH & 3,138  &1,046 &\underline{1,045} \\
    \midrule
        GSM8k~\cite{cobbe2021gsm8k} &Math Word Problems &EN &7,473 & - &\underline{1,319} \\
        AQuA-RAT~\cite{ling2017program} &Algebraic Word Problems &EN &97,467 &254 &\underline{254} \\
    \midrule
        % HumanEval~\cite{chen2021evaluating}  &Code Generation &EN & -& -&\underline{164} \\
    % \midrule
        MMLU~\cite{hendryckstest2021} &Complex Exams &EN &  -& 1,540 & \underline{14,049}\\
        C\_EVAL~\cite{huang2023ceval} &Complex Exams &ZH & - & 1,346 / 260 &\underline{12,342} \\
        GaoKao~\cite{zhong2023agieval} &Complex Exams &ZH / EN & -& -&\underline{2,080} \\
    \midrule
        AlpacaEval~\cite{alpaca_eval} &Alignment Evaluation &EN &  -& - & \underline{805}\\
        AlignBench~\cite{liu2023alignbench} &Alignment Evaluation &ZH & - & - &\underline{683} \\
    \bottomrule
    \end{tabular}}
    \label{tab:benchmarks}
\end{table*}

\paragraph{Construction of Training Dataset}
To support the human alignment initiative, we aggregate multiple datasets containing English and Chinese prompts alongside corresponding human preference data, which includes designated positive and negative responses. These datasets include HH-RLHF~\cite{hh_rlhf}, Stanford SHP~\cite{stanford_shp}, BeaverTails~\cite{beavertails}, Synthetic GPT-j,\footnote{\url{https://huggingface.co/datasets/Dahoas/synthetic-instruct-gptj-pairwise}} and UltraFeedback~\cite{cui2023ultrafeedback}, as well as the Chinese dataset CValues~\cite{xu2023cvalues}.
To enhance the reliability of the training data---ensuring the selected positive responses are decidedly superior to the negative---we apply a filtering mechanism based on user agreement counts in datasets such as Stanford SHP and BeaverTails, excluding any data where the disparity in agreement between responses falls below a predefined threshold.
% These datasets contain the prompt and corresponding human preference data, including positive response and negative response for the given prompt. To improve the quality of training data, \ie higher confidence that the positive response is strictly better than the negative response, we leverage the number of user agreements to filter the training data in Stanford SHP and BeaverTails, by removing the difference of agreements between positive response and negative response below a pre-defined threshold.

\paragraph{Difficulty Estimation based on DPO}
For human alignment, we use the DPO to fine-tune the model parameters. DPO evaluates the model's current capability against its counterpart before human alignment by comparing the discriminative power over positive and negative examples within each instance.  The reward calculation is formalized as:
% DPO directly compares the current model with its counterpart before human alignment to calculate the reward value, which directly assesses the discriminative degree of the current LLM for the contained positive and negative example in an instance. Thus, we can use its reward score to measure the difficulty of an instance as follows:
\begin{align}
R(p, y^+, y^-) =& \log(\frac{\pi_\theta(y^+|p)}{\pi_{\theta_{{o}}}(y^+|p)}) \notag\\
&- \log(\frac{\pi_\theta(y^-|p)}{\pi_{\theta_{{o}}}(y^-|p)}),  
\end{align}
where $\pi_\theta(y|p)$ and $\pi_{\theta_{o}}(y|p)$ denote the output distributions of the LLM trained after and before the current curriculum, respectively.  A higher reward value indicates that the model has effectively differentiated between the positive and negative examples, suggesting an increase in alignment accuracy. Conversely, lower reward values indicate the need for further learning, so we retain the corresponding data in subsequent training stages by applying a reward threshold $\delta$.
% Based on it, if the LLM has well learn to distinguish the positive or negative example during alignment training, the values of $\pi_\theta(y^+|p)$ and $\pi_\theta(y^-|p)$ will increase and decrease, respectively, resulting in an increased reward value. Conversely, if not, the reward value will become smaller. Then, by setting a threshold $\delta$, we can retain the instances with smaller rewards, which can be included in the curriculum of the next stage for further learning.

\paragraph{Easy-to-Hard Curriculum}
Based on the reward function , we can select and include challenging instances that the model has yet to master effectively. These are included in the subsequent training phases, setting a progressively decreasing threshold $\delta$ to increase the difficulty level. This approach follows the idea of curriculum learning where the model iteratively trains on increasingly challenging data. We optimize model parameters via the DPO strategy akin to fine-tuning, and the training objective is formulated as:
% In this way, we optimize the model parameters using DPO on the data from the current curriculum, which follows a strategy similar to fine-tuning, without the need for other reward models or complex reinforcement learning framework. The objective function is as follows:
\begin{align}
\nabla_\theta L_{DPO} = -\beta E_{(p, y^+, y^-) \sim \mathcal{D}} \sigma(R(p, y^+, y^-)) \notag \\ 
[\nabla \log{\pi\theta}(y^+|p) - \nabla \log{\pi_\theta}(y^-|p)], \notag
\end{align}
where $\beta$ is a hyper-parameter. The training focuses on maximizing the likelihood of generating value-aligned positive responses ($y^+$) and minimizing that of negative outputs ($y^-$). This learning process, which transitions from easier to more hard scenarios, ensures the model incrementally aligns closer to human preferences.
% During training, we would increase $\log{\pi_\theta}(y^+|p)$ and decrease $\log{\pi_\theta}(y^-|p)$, training the model to produce positive outputs $y^+$ that align with human values and avoiding generating the negative ones $y^-$. Throughout the iteration process, the easy-to-difficult curriculum helps the LLM selectively reinforce its learning of difficult instances, enabling it to gradually reach the human-aligned target.

\begin{table*}[t]
    \centering
    \small
    \caption{Zero-shot performance on commonsense reasoning benchmarks.}
    \setlength\tabcolsep{1mm}{
    \begin{tabular}{lccccccccc}
    \toprule
    & \textbf{BoolQ} & \textbf{PIQA} & \textbf{HellaSwag} & \textbf{WinoGrande} & \textbf{WSC273} & \textbf{ARC-e} & \textbf{ARC-c} & \textbf{OBQA} & \textbf{CommonsenseQA} \\
    \midrule
    Moss-moon-003-sft & 59.9  & 72.3  & 60.0  & 60.7  & 76.2  & 64.4  & 34.6  & 44.0  & 28.8  \\
    ChatGLM2 & 78.5  & 72.0  & 58.5  & 59.4  & 79.5  & 67.0  & 39.5  & 40.4  & 69.2 \\
    Baichuan2-13B & 78.7 & 77.4 & 78.6 & 75.4 & 87.2 & 77.4 & 71.5 & 47.6 & 69.7 \\
    Baichuan2-13B-chat & 81.4 & 75.4 & 77.4 & 76.1 & 84.6 & 75.2 & 68.8 & 47.0 & 71.0 \\
    LLaMA-13B & 76.4 & 79.7 & 79.7 & 79.5 & 90.5 & 77.4 & 68.2 & 47.2 & 63.6 \\
    LLaMA2-13B & 80.3 & 79.4 & 80.7 & 79.8 & 88.3 & 79.0 & 71.4 & 46.5 & 71.0 \\
    LLaMA2-13B-chat & 75.4  & 78.4  & 80.0  & 72.5  & 87.9  & 76.9  & 46.2  & 54.2  & 73.1 \\
    \midrule
    YuLan-Base & 69.1 & 76.1 & 72.3 & 65.8 & 85.4 & 71.8 & 41.0 & 52.6 & 59.4 \\
    YuLan-Inst & 79.8 & 77.5 & 74.6 & 71.7 & 81.7 & 77.6 & 49.5 & 57.4 & 77.3 \\
    YuLan-Chat & 83.5 & 78.1 & 76.7 & 72.7 & 83.2 & 80.1 & 51.6 & 58.4 & 76.7 \\
    \bottomrule
    \end{tabular}}
    \label{tab:reasoning}
\end{table*}

\begin{table*}[t]
    \centering
    \small
    \caption{The performance on factual knowledge, reading comprehension, and mathmatical reasoning benchmarks.}
    \setlength\tabcolsep{1mm}{
    \begin{tabular}{lccccccccc}
    \toprule
    & \textbf{TriviaQA}	& \textbf{RACE-m} & \textbf{RACE-h}	& \textbf{CoQA} & \textbf{CMRC2018} & \textbf{C3-Dialog} & \textbf{C3-Mixed} & \textbf{GSM8K} & \textbf{AQuA} \\
    \midrule
    Moss-moon-003-sft & 26.4  & 47.5  & 41.2  & 49.4  & 61.6  & 38.6  & 40.7  & 4.5  & 19.3 \\
    ChatGLM2 & 31.1  & 50.8  & 42.3  & 61.3  & 69.0  & 68.6  & 74.2  & 23.9  & 29.5 \\
    Baichuan2-13B & 66.2 & 55.0 & 46.4 & 80.3 & 74.2 & 81.0 & 76.5 & 42.8 & 35.8 \\
    Baichuan2-13B-chat & 65.1 & 60.5 & 54.1 & 75.5 & 77.1 & 86.8 & 86.3 & 46.3 & 34.7 \\
    LLaMA-13B & 73.9 & 49.2 & 45.7 & 77.5 & 66.2 & 41.5 & 44.8 & 17.1 & 19.7 \\
    LLaMA2-13B & 75.9 & 50.0 & 47.7 & 79.0 & 73.4 & 61.1 & 66.5 & 25.6 & 22.8 \\
    LLaMA2-13B-chat & 71.4  & 57.2  & 53.4  & 79.3  & 72.9  & 49.4  & 44.4  & 36.2  & 24.4 \\
    \midrule
    YuLan-Base & 59.3 & 48.3 & 43.0 & 77.2 & 64.9 & 47.1 & 45.2 & 18.6 & 15.8 \\
    YuLan-Inst & 47.9 & 49.7 & 45.2 & 77.0 & 73.3 & 83.8 & 82.5 & 29.6 & 28.7 \\
    YuLan-Chat & 45.6 & 56.6 & 47.2 & 71.4 & 72.4 & 83.2 & 82.6 & 30.1 & 27.2 \\
    \bottomrule
    \end{tabular}}
    \label{tab:reading}
\end{table*}

\section{Evaluation}
We evaluate our model on different NLP datasets and popular benchmarks, including commonsense and world knowledge, reading comprehension, math, code, and complex exams, as shown in Table~\ref{tab:benchmarks}. All the datasets and benchmarks can be split into two types: (1) classification problems, in which we need to calculate and compare the logits of different choices, such as BoolQ, PIQA, and MMLU; (2) generation problems, in which we need to extract and judge the answers from the generated contents, such as GSM8K, CommonsenseQA, and AQuA. We use the greedy decoding for the generation problem.

\subsection{Commonsense Reasoning}

We have carefully selected eight common datasets to assess the common sense reasoning capabilities of our models: BoolQ~\cite{clark2019boolq}, PIQA~\cite{Bisk2020PIQA}, Hellaswag~\cite{zellers2019hellaswag}, WinoGrande~\cite{ai2:winogrande}, WSC273~\cite{levesque2012winograd}, AI2\_ARC~\cite{allenai:arc}, OpenBookQA~\cite{OpenBookQA2018}, and CommonsenseQA~\cite{talmor-etal-2019-commonsenseqa}. Among these, the first seven datasets are treated as classification problems, and we adopt a zero-shot setting for evaluation. 
% Regarding CommonsenseQA, although it comprises multiple-choice questions, we approach it as a generation problem. 
We employ chain-of-thought methods, utilizing 7-shot examples for inference (consistent with~\cite{cot}). The experimental results are shown in Table~\ref{tab:reasoning}. We can observe that YuLan can achieve the best performance on BoolQ, OBQA, and CommonsenseQA, demonstrating its superior reaonsing capability. Besides, the instruction-tuning can significantly improves YuLan's performance (YuLan-Inst > YuLan-Base). This validates the effectiveness of our proposed instruction-tuning strategy.

\begin{table*}[t]
    \centering
    \small
    \caption{The performance on MMLU and C-Eval benchmarks.}
    \setlength\tabcolsep{0.8mm}{
    \begin{tabular}{lccccccccccc}
    \toprule
    & \multicolumn{5}{c}{MMLU} & \multicolumn{6}{c}{C-Eval} \\
    \cmidrule(r){2-6}\cmidrule(l){7-12}
     & \textbf{STEM} & \textbf{Social} & \textbf{Human} & \textbf{Other} & \textbf{Average} & \textbf{STEM} & \textbf{Social} & \textbf{Human} & \textbf{Other} & \textbf{Average} & \textbf{Hard} \\
     \midrule
     Moss-moon-003-sft & 27.2  & 29.1  & 29.5  & 32.8  & 29.6  & 30.0  & 36.0  & 33.1  & 32.0  & 32.2  & 26.7 \\
    ChatGLM2 & 38.9  & 52.3  & 43.0  & 52.2  & 46.6 & 46.5  & 65.4  & 52.4  & 47.6  & 51.6  & 33.5 \\
    Baichaun2-13B & 49.2 & 68.8 & 55.1 & 65.8 & 59.7 & 51.1 & 72.0 & 61.7 & 55.7 & 58.3 & 37.9 \\
    Baichaun2-13B-chat & 46.6 & 65.1 & 53.2 & 64.2 & 57.3 & 49.0 & 69.9 & 60.2 & 54.5 & 56.5 & 35.4 \\
    LLaMA-13B & 36.4 & 53.5 & 44.0 & 53.3 & 46.8 & 29.7 & 36.1 & 28.8 & 29.0 & 30.6 & 26.9 \\
    LLaMA2-13B & 44.6 & 64.2 & 53.9 & 62.2 & 56.2 & 36.9 & 43.2 & 37.6 & 36.6 & 38.2 & 32.0 \\
    LLaMA2-13B-chat & 40.4  & 60.6  & 45.2  & 57.5  & 50.9 & 33.6  & 40.9  & 34.1  & 35.6  & 35.5  & 27.3 \\
    \midrule
    YuLan-Base & 42.3 & 60.2 & 46.4 & 56.1 & 51.3 & 42.0 & 57.6 & 47.2 & 41.5 & 46.0 & 32.6 \\
    YuLan-Inst & 45.2 & 65.3 & 51.2 & 61.6 & 55.8 & 47.2 & 60.5 & 53.0 & 44.1 & 50.3 & 38.1 \\
    YuLan-Chat & 45.5 & 64.3 & 51.8 & 61.3 & 55.7 & 47.0 & 61.8 & 52.9 & 44.3 & 50.5 & 37.7 \\
    \bottomrule
    \end{tabular}}
    \label{tab:mmlu}
\end{table*}

\begin{table*}[t]
    \centering
    \small
    \caption{The performance on AGI-Gaokao tasks.}
    \setlength\tabcolsep{0.8mm}{
    \begin{tabular}{lcccccccccc}
    \toprule
     & \textbf{Chinese} &  \textbf{Geography} &  \textbf{Chemistry} &  \textbf{Biology} &  \textbf{Mathematics} &  \textbf{History} &  \textbf{English}  & \textbf{Physics} & \textbf{Average} \\
     \midrule
     Moss-moon-003-sft & 28.5  & 30.2  & 30.4  & 22.4  & 25.4  & 33.6  & 44.8  & 25.5  & 30.1 \\
    ChatGLM2 & 50.0  & 58.3  & 47.8  & 68.6  & 28.5  & 71.1  & 70.3  & 39.0  & 54.2 \\
    Baichaun2-13B & 50.4 & 68.8 & 43.5 & 58.6 & 31.9 & 70.6 & 78.8 & 33.0 & 54.5 \\
    Baichaun2-13B-chat & 48.4 & 65.8 & 44.4 & 57.6 & 31.1 & 67.7 & 78.4 & 28.5 & 52.7 \\
    LLaMA-13B & 22.8 & 26.6 & 31.4 & 23.8 & 26.2 & 23.8 & 58.8 & 26.5 & 30.0 \\
    LLaMA2-13B & 27.2 & 36.2 & 32.4 & 26.2 & 26.2 & 43.0 & 72.2 & 30.0 & 36.7 \\
    LLaMA2-13B-chat & 27.6  & 25.6  & 33.3  & 26.7  & 26.5  & 29.8  & 46.4  & 25.5  & 30.2 \\
    \midrule
    YuLan-Base & 31.3 & 53.3 & 34.8 & 43.8 & 28.2 & 60.9 & 68.3 & 27.5 & 43.5 \\
    YuLan-Inst & 42.3 & 57.3 & 41.6 & 54.3 & 27.9 & 68.5 & 80.4 & 25.5 & 49.7 \\
    YuLan-Chat & 43.9 & 57.3 & 37.7 & 53.8 & 26.2 & 69.4 & 80.4 & 27.0 & 49.5 \\
    \bottomrule
    \end{tabular}}
    \label{tab:gaokao}
\end{table*}

\subsection{Factual Knowledge}
% TriviaQA
We assess the factual knowledge within our models using the TriviaQA~\cite{2017arXivtriviaqa} dataset. We regard it as a generation problem and judge if the models' responses are contained in candidate answers. The evaluation results are shown in Table~\ref{tab:reading}. Unfortunately, there is still a gap between YuLan and other advanced LLMs. We attribute this to the gaps in data quality. Besides, we can see YuLan-Base performs better than YuLan-Inst and YuLan-Chat, reflecting instruction tuning and human alignment may affect LLMs' utilization of knowledge. However, more experiments are needed to explore the underneath reason. 

\subsection{Reading Comprehension}
% C3-dialog/mixed 
We evaluate the reading comprehension ability of our models using four widely used datasets: RACE~\cite{lai-etal-2017-race}, CoQA~\cite{reddy-etal-2019-coqa}, CMRC2018~\cite{cui-emnlp2019-cmrc2018}, and C3~\cite{sun2019investigating}. The latter two are Chinese datasets.
RACE and C3 are classification problems, while CoQA and CMRC2018 are generation problems. The evaluation results are shown in Table~\ref{tab:reading}. We can observe that YuLan, ChatGLM, and Baichuan can achieve significantly better performance on Chinese datasets, highlighting the importance of involving Chinese data in training process. Interestingly, LLaMA can perform well on CMRC, which is also a Chinese dataset. We check the dataset and find that it requires LLMs to select correct sentences from a provided document to answer the question. This capability may be easily transferred from learning on English data.

\begin{table*}[t]
\centering
\small
\caption{Comparison of different LLMs on AlpacaEval.}
\label{tab-yulan-alpacaeval}
\begin{tabular}{lcccc}
\toprule
 & \textbf{InternLM-7B-Chat} & \textbf{ChatGLM-6B} & \textbf{Baichuan-13B-Chat} & \textbf{MOSS-moon-003}  \\
 \midrule
\textbf{Win Ratio of YuLan} &  65.13\%  &   60.81\% & 59.57\%  & 57.06\%    \\
\bottomrule
\end{tabular}
\end{table*}

\begin{table}[t]
\centering
\small
\caption{Comparison of different LLMs on AlignBench. ``ZH'' denotes Chinese.}
\label{tab-yulan-alignbench}
\setlength{\tabcolsep}{0.8mm}{
\begin{tabular}{lccc}
\toprule
 & \textbf{ZH-Reasoning} & \textbf{ZH-Language} & \textbf{Avg.} \\
 \midrule
{InternLM-7B-Chat}        &  2.09 & 4.39 & 3.24   \\
{MOSS-moon-003}    & 2.24 & 4.67 & 3.46    \\
{ChatGLM-6B}                &  2.50 & 5.31 & 3.90   \\
{Baichuan-13B-Chat}       &  3.40 & 6.35 & 4.88   \\
\midrule
{YuLan-12B}                   & \textbf{3.59} & \textbf{6.69} & \textbf{5.14} \\
\bottomrule
\end{tabular}}
\end{table}

\subsection{Mathematical Reasoning}

% GSM8K 
% AQuA
We evaluate our models on two mathematical reasoning datasets: GSM8K~\cite{cobbe2021gsm8k} and AQuA-RAT~\cite{ling2017program}. We use their test sets ($1,319$ samples for GSM8k and $254$ samples for AQuA-RAT) with chain-of-thought examples (consistent with~\cite{cot}) for our evaluation. We regard them as generation problems and extract models' answers by regular expression. The experimental results are shown in Table~\ref{tab:reading}. We can see YuLan achieves comparable performance with LLaMA on these two datasets, indicating its capability of solving complex questions with chain-of-thought prompt. In our case study, we also test YuLan's capability of solving math problems in Chinese Gaokao.

% \subsection{Code Generation}
% % Human Eval
% We evaluate the code generation ability on HumanEval~\cite{chen2021evaluating}. This dataset includes $164$ programming problems with a function signature, doc string, body, and several unit tests. We use the zero-shot setting for our evaluation. 

\subsection{Comprehensive Benchmarks}

% MMLU
% C_EVAL
% GaoKao

We evaluate the abilities of our models to solve complex exams by MMLU~\cite{hendryckstest2021}, C-Eval~\cite{huang2023ceval}, and GaoKao. We use the $5$-shot examples as prompts for these benchmarks, the n-shot examples are provided by the benchmarks themselves (validation set). The majority of test samples in the three benchmarks consist of multiple-choice problems. As such, we treat them as classification problems, comparing the probabilities of various choices. The sole exception is the GaoKao-Math-Cloze task, which we regard as a generation problem due to the absence of candidate choices. The evaluation results are provided in Table~\ref{tab:mmlu} and Table~\ref{tab:gaokao}. Overall, YuLan achieves comparable performance with several advanced LLMs, demonstrating its capability of using acquired knowledge for solving real problems.

\subsection{Alignment Benchmarks}
We select two commonly-used benchmarks AlpacaEval~\cite{alpaca_eval} and AlignBench~\cite{liu2023alignbench} for the evaluation of alignment in LLMs. 
AlpacaEval is an English evaluation benchmark for human alignment, which utilizes powerful LLMs (\ie GPT-4) to perform pairwise comparisons of the outputs from two LLMs. Our analysis includes a comparison of our model, YuLan-Chat, against other baseline models. AlignBench is a Chinese benchmark, which performs multi-dimensional evaluation using chain-of-thought reasoning prompts to evaluate the models' responses comprehensively.

Table~\ref{tab-yulan-alpacaeval} shows the win rates of YuLan-Chat compared to other baseline models on AlpacaEval. YuLan-Chat demonstrates a win rate exceeding 55\%, indicating its superior alignment with human preferences. This improvement is largely due to its curriculum-based instruction tuning and human alignment training strategies, which facilitate the model's comprehension of complex instructions and generation of unbiased responses. Table~\ref{tab-yulan-alignbench} shows the performance of various LLMs on AlignBench, with a specific focus on Chinese language alignment. Notably, the alignment capabilities in Chinese and English across these models are quite different. Among the baselines, Baichuan-13B-Chat performs optimally in Chinese alignment, benefiting from extensive use of human-annotated data tailored for this purpose. Additionally, YuLan-Chat surpasses all baseline models, attributed to its multi-stage fine-tuning through curriculum learning. This training method not only improve performance on intricate reasoning tasks but also ensures robustness in Chinese linguistic proficiency.

% \begin{table}[t]
%     \centering
%     \small
%     \setlength\tabcolsep{0.6mm}{
%     \begin{tabular}{lcccccc}
%     \toprule
%      & STEM & Social & Human & Other & Average & Hard \\
%      \midrule
%      Moss-moon-003 \\
%     ChatGLM2  \\
%     Baichaun2  \\
%     Baichaun2-chat  \\
%     LLaMA-13B  \\
%     LLaMA2-13B  \\
%     LLaMA2-chat  \\
%     \midrule
%     YuLan-Base  \\
%     YuLan-Inst  \\
%     YuLan-Chat  \\
%     \bottomrule
%     \end{tabular}}
%     \caption{The performance on C-Eval tasks.}
%     \label{tab:benchmarks}
% \end{table}

\section{Discussion}
We also conduct a series of experiments to validate some strategies in the training of YuLan.

\begin{table}[t]
    \centering
    \small
    \caption{Comparison of using different strategies in pre-training.}
    \setlength{\tabcolsep}{1.5mm}{
    \begin{tabular}{lccc}
    \toprule
        \textbf{Benchmark} & \textbf{Strategy 1} & \textbf{Strategy 2} & \textbf{Strategy 3} \\
    \midrule
        CommonsenseQA & 13.68 & 15.64 & \textbf{18.43} \\
        AQuA & 13.78 & 13.78 & \textbf{15.75} \\
        CMRC (Chinese) & 12.99 & \textbf{17.61} & 16.41 \\
        MMLU & 25.83 & \textbf{43.30} & 43.15 \\
        AGI-Gaokao & 27.15 & \textbf{40.71} & 38.81 \\
    \bottomrule
    \end{tabular}
    }
    \label{tab:strategy}
\end{table}

\begin{table}[!t]
\centering
\small
\caption{Effect of different Chinese data categories and distributions (dist.) on model performance.}
% \scalebox{0.95}{
\begin{tabular}{lcccc}
\toprule
\textbf{Data} & \textbf{Dist.} & \textbf{MMLU}          & \textbf{C-EVAL}        & \textbf{(Hard)} \\ \midrule
HQ & 1            & 26.6          & 24.5          & 23.9          \\
Web & 1            & 24.9          & 25.1          & 25.3          \\
News & 1            & 25.7          & 25.6          & 27.3          \\
Law & 1            & 25.1          & \textbf{26.6} & \textbf{28.1} \\
HQ+Web & 4:1          & 25.7          & 25.6          & 26.0          \\
HQ+News & 4:1          & 26.5          & 25.6          & 26.2          \\
HQ+Law & 4:1          & 24.1          & 24.9          & 24.1          \\
{HQ+All} & 4:2:2:2      & \textbf{26.8} & 26.1          & 25.7          \\ \bottomrule
\end{tabular}
% }
\label{table:effect_of_chinese_data_sources}
\end{table}

\begin{table*}[!t]
\small
\centering
\caption{Performance comparisons of different training strategies for investigating the effect of continual pre-training.}
\begin{tabular}{ccccc}
\toprule
\textbf{Model} & \textbf{Data}             & \textbf{MMLU} & \textbf{C-EVAL} & {\textbf{C-EVAL (Hard)}} \\ \midrule
\textit{v1}    & 20B EN + 10B ZH ($p_1$ + $p_2$)     & 26.8 & \textbf{26.1}   & 25.7          \\
\textit{v2}    & 13B EN + 6.5B ZH ($p_1$) & 26.4 & 24.1   & 25.4          \\
\textit{v3}    & 7B EN + 3.5B ZH ($p_2$)  & \textbf{27.3} & \textbf{26.1}   & \textbf{27.3}          \\ \bottomrule
\end{tabular}
\label{table:effect_of_continue_pretraining}
\end{table*}

\subsection{Impact of Educational Assessments}\label{sec:exam}
During the first pre-training stage, we observe fluctuations in model's performance, particularly with comprehensive benchmarks such as MMLU. To address this problem, we conduct three experiments by applying different strategies to continue pre-training $5,000$ steps based on the $600$B checkpoint: 

\noindent$\bullet$ Strategy 1: Maintaining the original data distribution. 

\noindent$\bullet$ Strategy 2: Adding educational assessments~(in Chinese). 

\noindent$\bullet$ Strategy 3: Adding both MMLU-related QA training sets and educational assessments~(in Chinese). 

The experimental results are shown in Table~\ref{tab:strategy}. In the first stage, we exclude educational assessments from our pre-training to avoid their unique formats potentially skewing model training. However, the absence of this specific form of knowledge appeared to limit the model's ability to contextualize and apply learned information effectively, as evidenced by the inferior performance in both the MMLU and AGI-Gaokao benchmarks under Strategy 1. Intriguingly, the inclusion of Chinese educational assessments in Strategy 2 substantially enhanced performance across both the English MMLU and Chinese AGI-Gaokao benchmarks, illustrating a cross-lingual transferability in the understanding and application of task-specific knowledge. In Strategy 3, the addition of MMLU-related QA training further elevated the model's performance on various QA benchmarks, corroborating the significant impact of diverse task forms on model enhancement. Given these findings, we opt to incorporate educational assessments into our pre-training dataset in the second pre-training stage, affirming the pivotal role of diverse educational content in enhancing language model performance.\footnote{To avoid data leakage~\cite{leakage}, we still avoid using MMLU-related training sets.}

\begin{figure*}
    \centering
    \includegraphics[width=\linewidth]{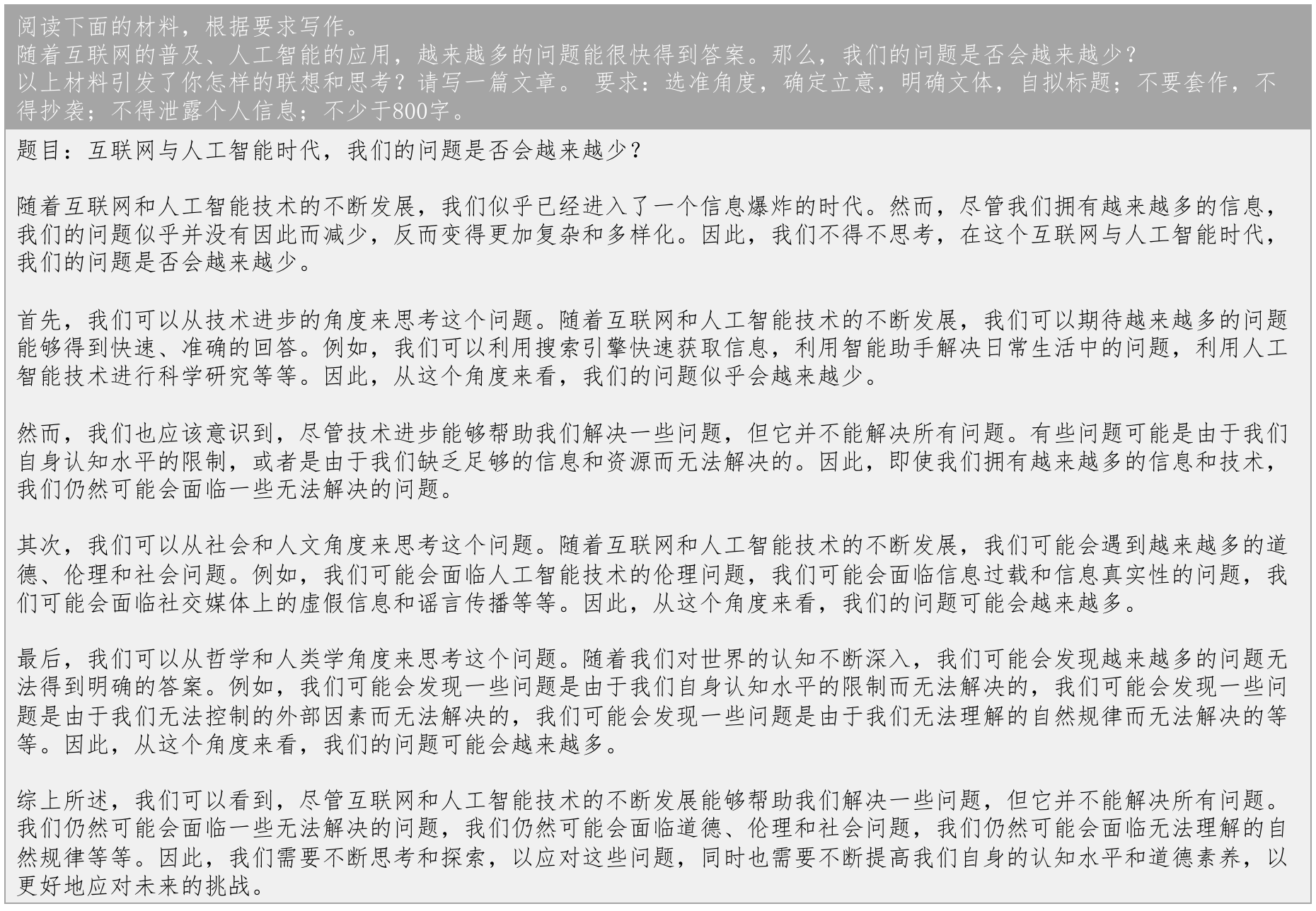}
    \caption{An example of YuLan-Chat on writing a 2024 Gaokao Chinese essay.}
    \label{fig:example1}
\end{figure*}

\subsection{Preliminary Experiments with 1.3B Model}

Before finalizing our training strategy, we conducted a series of preliminary experiments using the smaller 1.3B model to examine the impact of Chinese data, continual training, and longer context window size on model performance.

\paragraph{Effect of Chinese Data}
We explore the impact of various Chinese data types on model performance. The Chinese data is categorized into four types:
(1) High-quality (HQ), including books, QA forums (Zhihu), and Wikipedia,
(2) Web,
(3) News,
and (4) Law.
We randomly sample 10B Chinese tokens from different combinations of these categories and 20B English tokens based on the LLaMA data distribution. A series of 1.3B models are pre-trained on these different 30B tokens, and their performances are evaluated using MMLU, C-EVAL, and C-EVAL (Hard). The results are summarized in Table \ref{table:effect_of_chinese_data_sources}.
We find that:

(1) Mixing multiple sources of data can enhance performance for both English and Chinese. The model trained on a mix of all four Chinese data sources achieves the best performance on MMLU and the second best on C-EVAL.

(2) High-quality Chinese data can improve English performance but may not be sufficient for the best Chinese performance. Using high-quality data alone achieves the second-best MMLU score but performs worse than web, news, or law data alone on C-EVAL. Combining high-quality data with other sources significantly improves C-EVAL performance. This suggests that a diverse data mix is essential to balance English and Chinese performance, rather than relying solely on high-quality data from limited sources.

(3) Our law data exhibits an unusual pattern, achieving the highest scores on C-EVAL and C-EVAL (Hard) when used alone. However, mixing law data with high-quality data results in a significant performance drop, yielding the worst results for both English and Chinese tasks. Therefore, we decide to just mix a small amount of law data in our final pre-training.

\paragraph{Effect of Continual Training}
We explore the impact of continual training by investigating whether ``continual training with newly added data sources'' can achieve performance levels similar to ``training with all data sources from the beginning''. Specifically, we compare the following three settings:
(1) \textit{v1}: Training on $20$B English tokens and $10$B Chinese tokens together from the beginning.
Then, we divide the $10$B Chinese tokens into two parts: $p_1$ (books, news, wiki, totaling $6.5$B tokens) and $p_2$ (law, Zhihu, totaling $3.5$B tokens).
(2) \textit{v2}: We train on $13$B English tokens and $6.5$B Chinese tokens ($p_1$). 
(3) \textit{v3}: We continue training \textit{v2} on $7$B English tokens and $3.5$B Chinese tokens ($p_2$).
The results are shown in Table~\ref{table:effect_of_continue_pretraining}.
We find that continual training with newly added data sources yields better (or at least equivalent) performance in both English and Chinese compared to training with all data sources from the beginning. This demonstrates that continual training with new data sources is effective.

\begin{figure*}
    \centering
    \includegraphics[width=\linewidth]{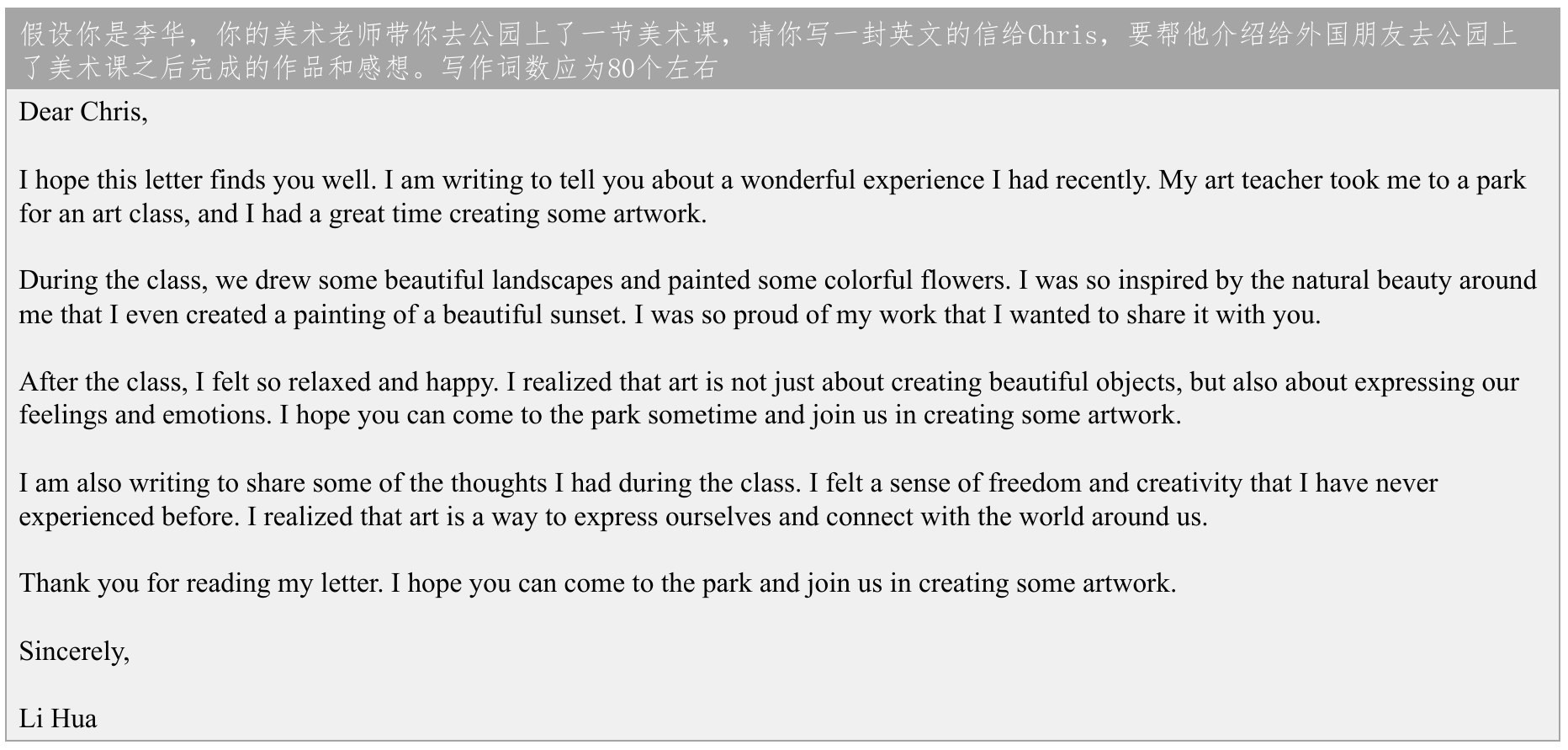}
    \caption{An example of YuLan-Chat on writing a 2024 Gaokao English essay.}
    \label{fig:example2}
\end{figure*}

\begin{figure*}
    \centering
    \includegraphics[width=\linewidth]{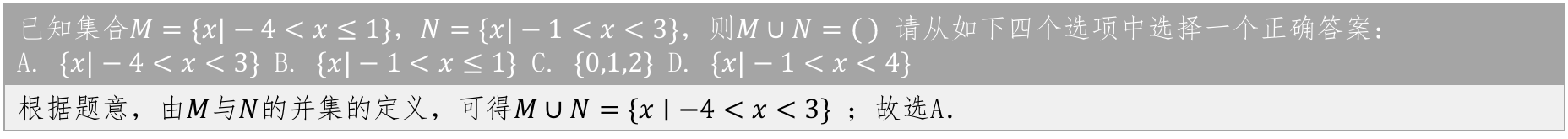}
    \caption{An example of YuLan-Chat on answering a 2024 Gaokao math question.}
    \label{fig:example3}
\end{figure*}

\begin{figure*}
    \centering
    \includegraphics[width=\linewidth]{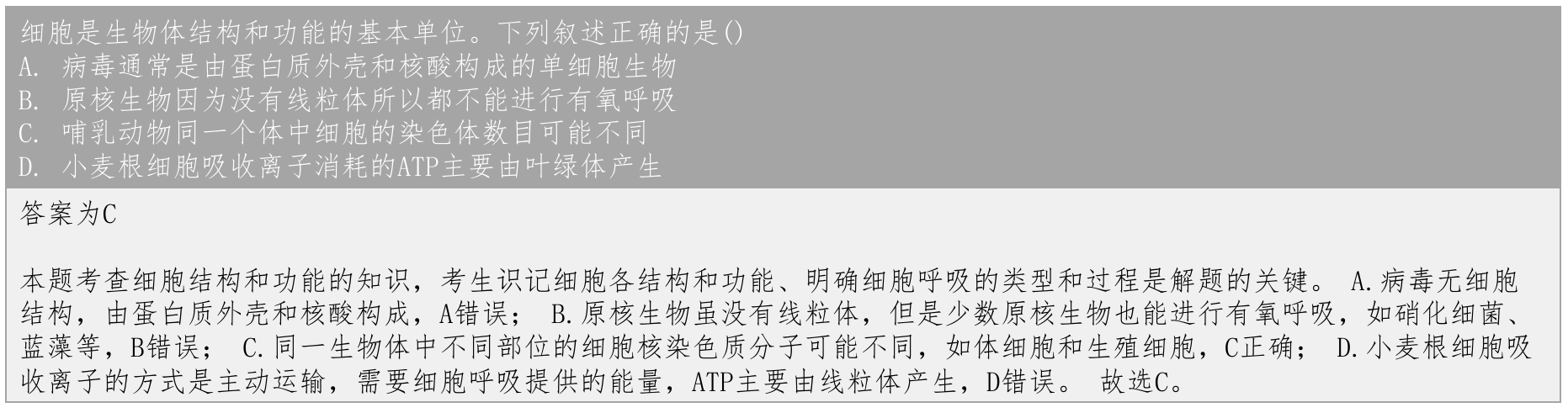}
    \caption{An example of YuLan-Chat on answering a 2024 Gaokao biology question.}
    \label{fig:example4}
\end{figure*}

\begin{figure*}
    \centering
    \includegraphics[width=\linewidth]{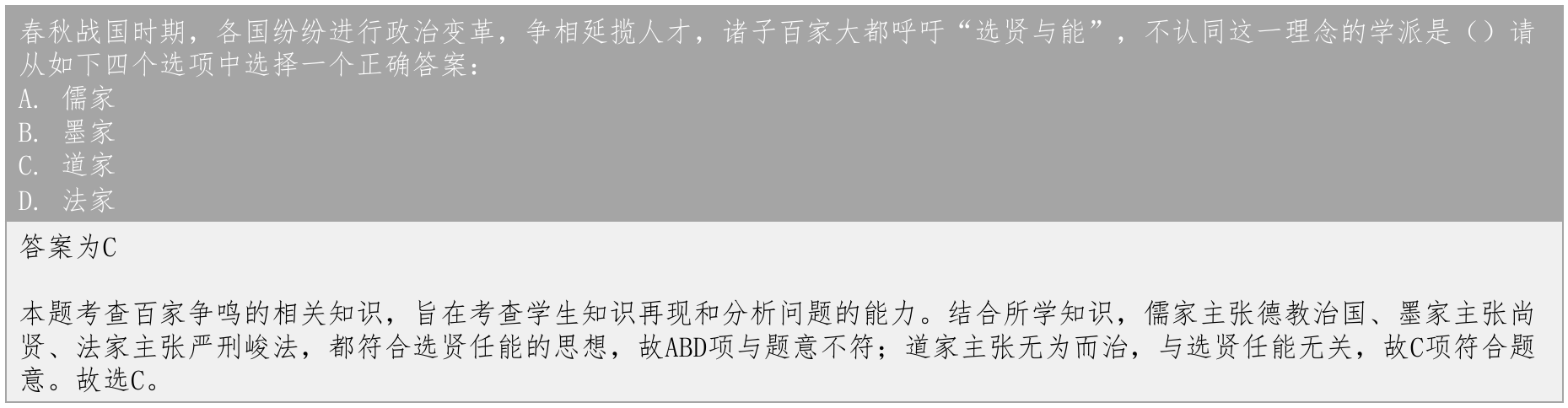}
    \caption{An example of YuLan-Chat on answering a 2024 Gaokao history question.}
    \label{fig:example5}
\end{figure*}

\begin{figure*}
    \centering
    \includegraphics[width=\linewidth]{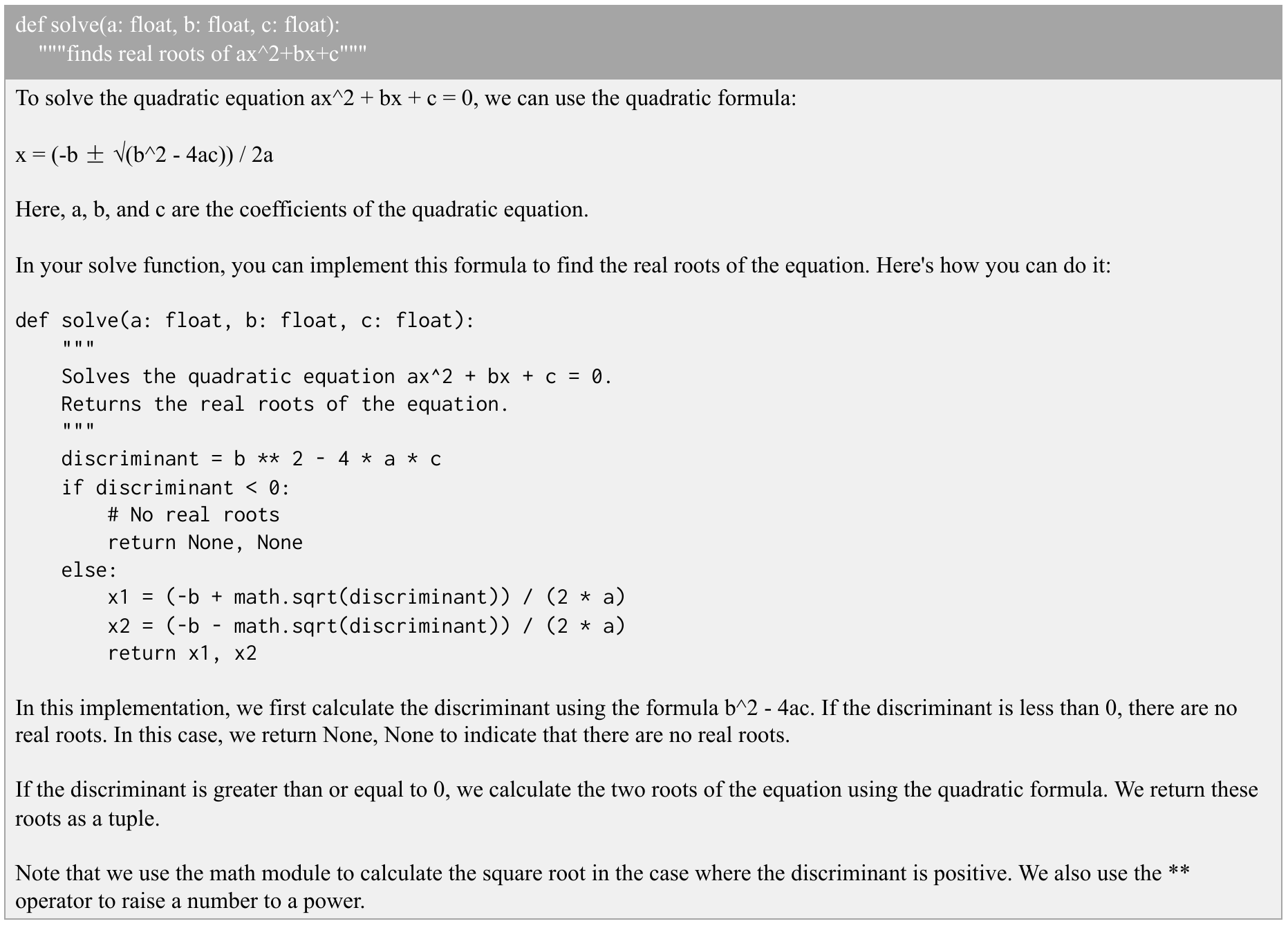}
    \caption{An example of YuLan-Chat on writing code.}
    \label{fig:example6}
\end{figure*}

\begin{figure*}
    \centering
    \includegraphics[width=\linewidth]{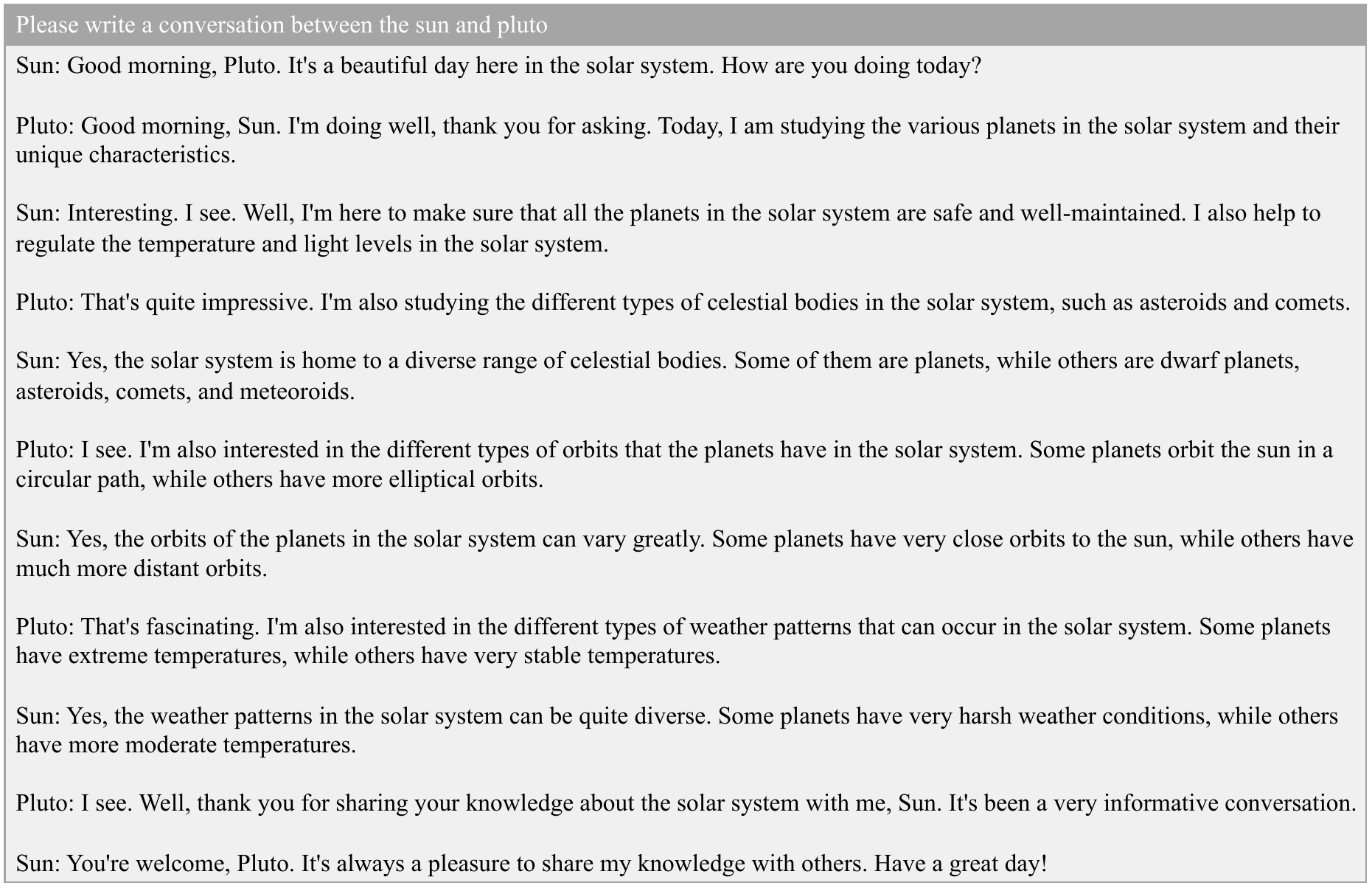}
    \caption{An example of YuLan-Chat on generating stories.}
    \label{fig:example7}
\end{figure*}

\begin{figure*}
    \centering
    \includegraphics[width=\linewidth]{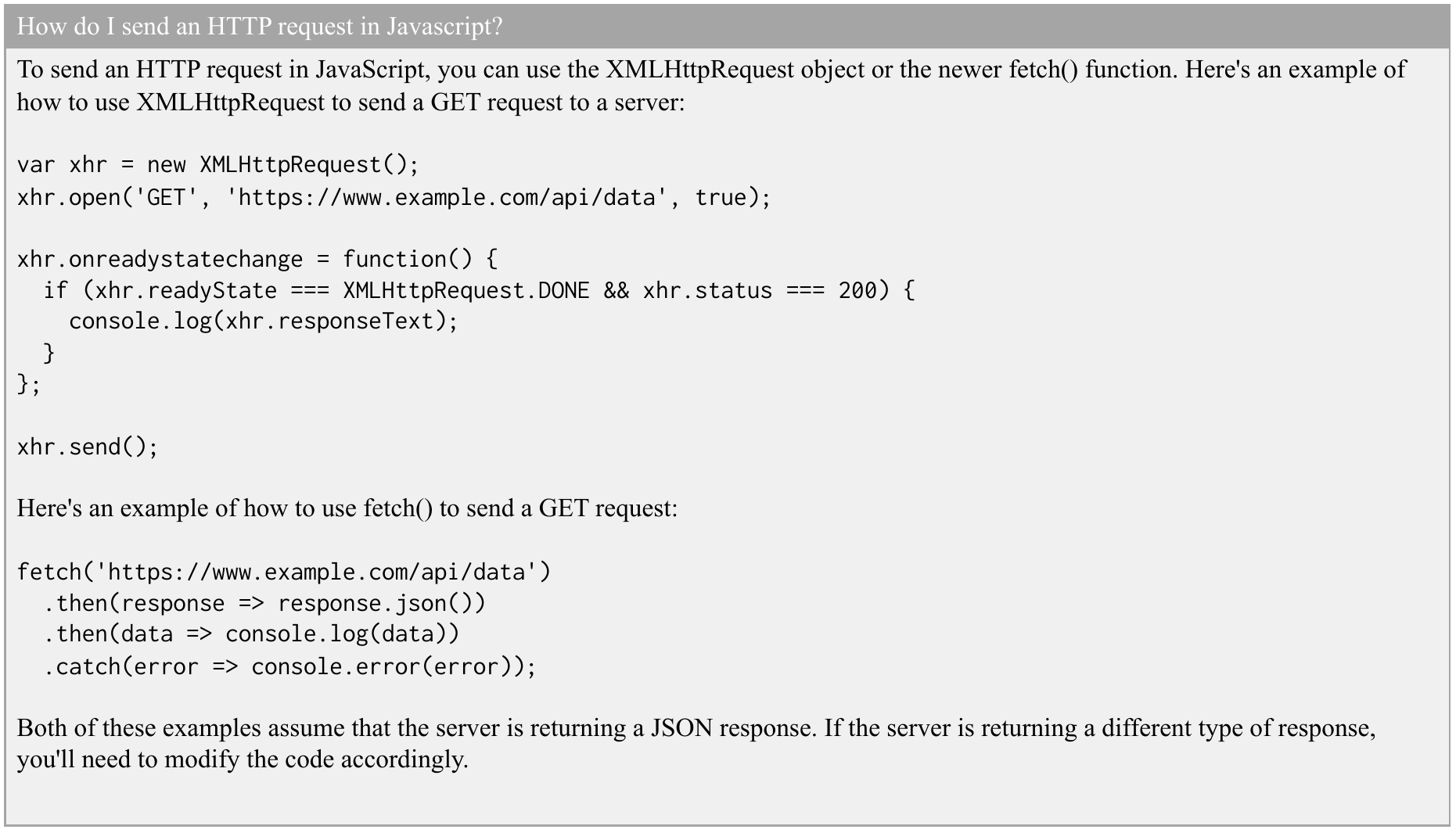}
    \caption{An example of YuLan-Chat on answering programming problems.}
    \label{fig:example8}
\end{figure*}

\subsection{Case Study}
We perform a case study by testing YuLan-Chat on a range of subjects from the 2024 Chinese Gaokao, encompassing Chinese essay writing, English essay writing, mathematics, biology, and history. This diversity of subjects provided a comprehensive assessment of YuLan's capabilities across different subjects. The generation results are shown in Figure~\ref{fig:example1}-\ref{fig:example5}. We can observe that YuLan-Chat excel in Chinese and English essay writing, demonstrating a strong ability to construct coherent, well-argued natural language texts. In mathematics, the model efficiently solves the problem, showcasing its quantitative reasoning skills. Similarly, in biology and history, YuLan-Chat accurately answers questions involving intricate details and conceptual understanding. Besides, we also show YuLan-Chat's ability on writing code, generating stories, and answering real problems in Figure~\ref{fig:example6}-\ref{fig:example8}. These results confirm YuLan-Chat's adaptability and intellectual breadth.

\section{Conclusion}
In this report, we introduced the detailed training process of YuLan-12B, including pre-training, supervised fine-tuning, and human alignment. The YuLan-12B, trained on approximately $1.7$TB tokens, has demonstrated performance on par with other open-source LLMs. Despite a surge in advanced models trained on larger datasets, this paper aims to illuminate essential training techniques for developing LLMs from scratch and to provide insights for future research. We hope our report can enhance understanding and foster innovation within the AI community, promoting transparency and reproducibility in AI research.
% By sharing our findings, we also contribute to democratizing AI knowledge, ensuring accessibility for researchers across diverse regions and institutions.

\section*{Acknowledgment}
The computing resources are supported by Public Computing Cloud, Renmin University of China.

% \clearpage

% Bibliography entries for the entire Anthology, followed by custom entries
%\bibliography{anthology,custom}
% Custom bibliography entries only
% \bibliography{main}

\balance
\appendix
\section{Detailed Data and Processing}\label{app:data}
\subsection{Web Pages}\label{sec.data.cc}
The Internet, as a comprehensive and continually updated source of information, provides rich contents that are invaluable for training LLMs. Web pages offer a broad spectrum of knowledge across various domains, making them essential for developing models that are robust and capable of understanding context in multiple fields. This diversity not only enriches the training set but also enhances the generalizability and applicability of LLMs in real-world scenarios. To leverage this vast resource, we have combined several key datasets.

Specifically, our dataset includes data from OpenWebText2~\cite{thepile}, C4~\cite{DBLP:conf/emnlp/DodgeSMAIGM021}, RefinedWeb~\cite{refinedweb}, CC-100~\cite{cc100}, ClueWeb 22~\cite{clueweb22}, CC-Stories~\cite{ccstories}, and Dolma's CC~\cite{dolma}. In addition to these sources, we process raw data from Common Crawl (CC) dumps, particularly focusing on events that occurred between January 2021 and February 2023. To manage the practical challenges of HTML content extraction and constraints of disk space, we utilize the WET file format, which includes only plain text, for further preprocessing. We selectively retain texts in English, Chinese, and other multilingual texts that are specified in our language list, ensuring a diverse yet controlled dataset for model training.

\paragraph{Preprocessing}
Generally, the preprocessing of data involves a three-stage procedure designed to enhance data quality significantly:
(1) We apply heuristic rules at both the coarse and fine-grained levels. This initial filtering focuses on the structural and content aspects of texts.
(2) We employ the CCNet pipeline~\cite{ccnet} to perform deduplication, identify language, and assess the linguistic quality of texts.
(3) We conduct a final quality check with several heuristic rules.

Specifically, in Stage~(1), we conduct filtering on the page and paragraph level. We exclude pages shorter than $512$ characters or those predominantly ($>50\%$) consisting of sentences shorter than $16$ characters due to their limited contextual value. Content containing non-relevant elements such as ``javascript'', ``lorem ipsum'', curly brackets ``\{'', or terms from a list of dirty words is removed to maintain the quality and relevance of the data.\footnote{\url{https://github.com/LDNOOBW/List-of-Dirty-Naughty-Obscene-and-Otherwise-Bad-Words}}
% \footnote{\href{https://github.com/LDNOOBW/List-of-Dirty-Naughty-Obscene-and-Otherwise-Bad-Words}{https://github.com/LDNOOBW/List-of-Dirty-Naughty-Obscene-and-Otherwise-Bad-Words}}
We also eliminate pages with a high presence of hash symbols or ellipsis~(ratio $>0.1$), and those heavily formatted with bullet points~($>90\%$ lines) or frequently ending in ellipses~($>30\%$ lines), as these features typically indicate poor quality or non-standard text formats. Finally, we ensure that retained texts end with proper punctuation, and we discard any text with garbled characters or unresolved Unicode conversions. In Stage~(2), we use the CCNet pipeline to evaluate the complexity and readability of the text, with pages having a perplexity score over 1,000 being discarded. We conduct language identification to ensure the text matches our target languages, retaining only those with a language score above $0.6$. In Stage~(3), we remove texts that are excessively short or repetitive. Specifically, pages shorter than $500$ characters or with a low total number of paragraphs~(less than three) indicating high repetitiveness are excluded. We also filter out texts where the original number of paragraphs is disproportionately high (more than five times) compared to the number retained after processing, ensuring content consistency and integrity.

% and language scores of each page, while deduplication is conducted at a paragraph level. 
% After being processed by CCNet, 
% in stage (3), we remove pages with a PPL score over $1,000$ and a length below $500$ characters.
% Since a smaller total number of paragraphs means a higher paragraph repetition ratio, we filter out those with a total number of paragraphs less than three. Besides, samples are discarded if the original total number of paragraphs is greater than five times the number of processed paragraphs. Finally, we only keep those texts with language scores greater than $0.6$.
% For the CC dataset we built on our own, we take some more strict rules like removing those pages with a length of less than 512 characters in stage (1).
% We directly utilize the RefinedWeb because of its relatively high quality, while the CC-stories dataset is used after removing dirty words.

\subsection{Code}
% Yiding Sun
Incorporating programming code into pre-training data is critical for enhancing the capabilities of LLMs, particularly in fostering the development of an emergent chain-of-thought and algorithmic reasoning. Code inherently embodies structured, logical thinking and provides a sequential understanding of tasks, which are fundamental to developing LLMs that can emulate human-like problem-solving skills. Studies have shown that the inclusion of programming code not only augments the syntactic understanding but also significantly boosts the model's ability to perform complex reasoning and execute task-specific functions~\cite{gpt3, zhao2023survey}. Hence, our dataset extensively incorporates code from various sources to cultivate these advanced capabilities in our LLM. We source the programming code from two primary repositories: the Stack~\cite{stack} and GitHub.

The Stack, part of the BigCode Project, holds over $6$TB of source code files across $358$ programming languages, emphasizing the broad spectrum of coding knowledge available. For the purpose of our model, we focus exclusively on Python due to its wide usage and relevance in both academic and practical applications. Given that the data from the Stack has already undergone preliminary processing to ensure consistency and quality, we do not perform additional preprocessing steps.

GitHub, as a vast open-source platform, hosts approximately $28$ million public repositories as of January 2023. This platform is a treasure trove of code across various programming languages, offering a real-world mix of codebases. However, the data from GitHub often contains a significant amount of non-code elements or noise. To enhance data quality, we apply a meticulous selection process, initially filtering repositories based on popularity (\eg{} those with over 100 stars as of March 12, 2023). Subsequently, we clone these repositories, removing non-code files and retaining markdown files, particularly READMEs, under the premise that these documents provide valuable context and explanations that aid in the model's deeper understanding of the code.

Following collection, we perform rigorous preprocessing to further refine the quality and relevancy of the code data, ensuring that our LLM is trained on high-quality, representative programming content that enhances its coding ability and logical reasoning skills.

\paragraph{Preprocessing}
Following previous studies~\cite{roots, llama, palm}, our preprocessing routine for programming code employs several heuristic rules designed to refine the quality and relevance of the data for training purposes. Initially, we filter out files based on several criteria. We eliminate files with fewer than $100$ characters, except for those ending in ``.sql'', or more than $200,000$ characters to maintain an optimal range of complexity and detail. Files with any line shorter than $20$ characters or longer than $1,000$ characters are discarded, as they often do not represent standard coding practices. We remove files where numeric characters exceed $70\%$ of the content or alphabetical characters constitute less than $30\%$, to avoid files dominated by data values or non-instructional content. Files are also excluded if they include exact matches to the phrases ``configuration file'' or ``test file'', or if over $10\%$ of the lines contain the words ``config'' or ``test'', indicating non-functional code such as configuration or test scripts. 
The majority of excluded files typically include data dumps, configuration files, log files, or automatically generated template code, which lack substantial logical coding content necessary for effective model training. 

Following the initial filtering phase, we proceed with deduplication. We first apply exact match filtering within predefined slices of the dataset to manage memory use efficiently. After this intra-slice deduplication, we consolidate all slices and apply a 10-gram minhashLSH algorithm to perform comprehensive deduplication across the entire dataset. This two-step deduplication process ensures that the final training dataset is devoid of redundant entries, thereby enhancing the quality and efficiency of the training phase.

\subsection{Encyclopedia}
% Wentong Chen
Encyclopedias represent a cornerstone resource in the pre-training of LLMs, offering a vast repository of structured, high-quality human knowledge essential for building comprehensive understanding. These resources are pivotal in enhancing the factual accuracy and depth of knowledge of LLMs, making them necessary for applications requiring reliable information and nuanced content generation. In our pre-training, we extend beyond the conventional use of Wikipedia to include the Baidu Encyclopedia, thereby enriching our dataset with expansive Chinese linguistic and cultural knowledge.

Wikipedia, launched by Jimmy Wales and Larry Sanger on January 15, 2001, stands as the most extensive free-content encyclopedia available online. It is maintained by a global volunteer community using a wiki-based editing system, MediaWiki, and is one of the top ten most visited websites globally. We gather data from Wikipedia in English, Chinese, and other languages in our multilingual list directly from the Wikimedia Downloads site.\footnote{\url{https://dumps.wikimedia.org/backup-index.html}}

In addition to Wikipedia, we incorporate data from the Baidu Encyclopedia, the largest semi-regulated online Chinese encyclopedia managed by Baidu, Inc. It allows user-contributed content, which is subsequently reviewed by official editors to ensure the accuracy and relevance of the information. As of July 2023, it encompasses nearly $27$ million entries in both Simplified and Traditional Chinese. Due to Baidu's strict anti-crawler policies, we access this encyclopedia through a third-party collection available on HuggingFace, initially collected in 2020.\footnote{\url{https://huggingface.co/datasets/TMZN/baidubaike}} To incorporate the most current entries, we also utilize a more recent version provided by Tiger Research.\footnote{\url{https://huggingface.co/datasets/TigerResearch/pretrain\_zh}}

By integrating these diverse encyclopedic sources, we aim to construct a robust LLM that is well-versed across multiple languages and domains, capable of generating accurate and culturally relevant content.

% \paragraph{Baidu Encyclopedia}

% \paragraph{Tiger Encyclopedia}

% Tiger Encyclopedia is the encyclopedic part of high-quality pre-training data collected by TigerResearch.
% It consists of well-formatted Chinese and English encyclopedic entries. 
% We filtered nearly 7 million Chinese entries to further enrich the diversity and abundance of our Encyclopedia.
\paragraph{Preprocessing} 
For Wikipedia, we begin with the extraction of content using the \texttt{wikiextractor} tool, which parses XML dumps to isolate meaningful textual data.\footnote{\url{https://www.cnpython.com/pypi/wikiextractor}} During this process, we preserve article titles and textual paragraphs while eliminating non-textual elements such as images, tables, audio, and video files. Subsequently, the extracted data is converted to a JSONL format for better handling and integration into our dataset. Each Wikipedia article is structured into a single line in this format, with the title and paragraphs separated by newline characters. For the Chinese Wikipedia, we use \texttt{zhconv} to standardize all text to simplified Chinese, ensuring consistency across our Chinese language data.\footnote{\url{https://www.cnpython.com/pypi/zhconv}}

For Baidu Encyclopedia, to enhance its quality, we implement a set of heuristic rules aimed at refining the content: (1) We strip all entries of meaningless headers and footers, which often contain repetitive or irrelevant information that may affect the quality of the training data. (2) We discard entries that are shorter than $50$ characters or have a proportion of Chinese characters below $70\%$, as such entries typically lack substantial informational content. (3) The remaining text from each entry is converted into simplified Chinese and recombined with the original titles to produce cohesive and standardized entries. 

These preprocessing steps are designed to ensure that the encyclopedia data fed into our model training pipeline is clean, uniform, and optimally formatted, thereby facilitating the development of a more effective and knowledgeable language model.

\subsection{Academic Papers}
% Yihan Wu
Academic papers are a pivotal source for the pre-training of LLMs due to their complex structure, formal language, and rich scientific content. These documents provide a diverse array of knowledge and are instrumental in enhancing the reasoning capabilities of LLMs, allowing them to perform more effectively in tasks requiring deep understanding and analytical skills. To this end, we incorporate a substantial corpus of papers from two major repositories: arXiv and the peS2o dataset~\cite{peS2o}.

% We process ArXiv Latex files to provide scientific knowledge to our dataset. 
ArXiv is an open-access archive that hosts over $2.3$ million scholarly articles spanning diverse scientific domains such as physics, mathematics, computer science, and economics, among others. We systematically collect all LaTeX files available from 1990 to March 2023 via the arXiv bulk data service on Amazon S3.\footnote{\url{https://info.arxiv.org/help/bulk_data_s3.html}} These documents are a rich source of advanced scientific and technical knowledge, ideal for training sophisticated LLMs.

The peS2o dataset contains approximately $40$ million open-access academic papers, derived from the S2ORC project~\cite{DBLP:conf/acl/LoWNKW20}. The peS2o dataset has undergone extensive preprocessing by the dataset creators, including comprehensive cleaning, filtering, and formatting, ensuring its readiness for integration into our training pipeline.

\paragraph{Preprocessing}
For the arXiv dataset, in line with established practices from prior studies~\cite{llama,thepile}, we convert LaTeX files into markdown format using \texttt{pandoc}. This transformation facilitates the removal of non-essential elements such as titles, author details, bibliographies, and any content preceding the introduction. Additionally, we standardize the format by normalizing multiple consecutive blank lines to a single blank line, enhancing the readability and consistency of the text for model training.

The integration of academic papers enriches our training dataset with formal, structured, and authoritative scientific discourse, significantly benefiting the cognitive and reasoning faculties of the resultant LLM. These elements are critical for applications that demand high levels of comprehension, analytical depth, and factual accuracy, such as academic research assistance, technical writing, and complex problem-solving.

\subsection{QA Forums}
Question-answering datasets are crucial for the pre-training of LLMs, as they provide the necessary supervisory signals for LLMs and promote the improvement of the models' capabilities in language understanding, knowledge acquisition, context awareness, generalization, and dialogue generation. The improvement of these capabilities is crucial for developing more intelligent, efficient, and practical LLMs. We use the Stack Exchange (in English) dataset and the Zhihu (in Chinese) dataset.

% \paragraph{Stack Exchange}
% Qiangqiang Ren
% The Stack Exchange dataset is a public data resource provided by the Stack Exchange Network, which contains data from multiple online question-and-answer communities. These communities include the following types: Technology, Culture \& recreation, Life \& arts, Science, Professional, and Business, which cover a variety of topics in daily work and study life, such as programming, mathematics, network, English language \& usage, games, operating systems, and more.

Stack Exchange is a network of question-and-answer websites on topics in diverse fields, each site covering a specific topic, where questions, answers, and users are subject to a reputation award process. Stack Exchange sites are designed to foster expert communities where users can ask questions and provide quality answers, receiving reputation points and badges as rewards for helpful contributions.

Zhihu is a Chinese question-and-answer forum that serves as a comprehensive platform for users to exchange knowledge, experiences, and insights. Users on Zhihu can pose questions on a vast array of topics, ranging from science and technology to culture and education, and receive answers from other community members. These responses can be upvoted or downvoted by users, allowing the most valuable content to be easily accessible.

\paragraph{Preprocessing}
The Stack Exchange Data Dump contains an anonymized set of all user-contributed content across the Stack Exchange network, organized into site-specific archives. Each archive is formatted as a zipped XML file and contains various data attributes including Posts, Users, Votes, Comments, Badges, Tags,  PostHistory, and PostLinks.\footnote{\url{https://archive.org/details/stackexchange}} 

Given the diverse quality of the question-and-answer data contained within these dumps, rigorous preprocessing is essential to obtain high-quality data. Following existing studies~\cite{llama,thepile}, we process the dataset in several steps. Initially, we parse the XML data to extract textual information from questions and answers, along with important metadata such as the Score attribute. This Score, which ranges from 0 to 10, indicates the quality of an answer---the higher the Score, the higher the quality. To assess the reliability of the Score as a quality indicator, we perform a statistical analysis across the distribution of scores and manually review some random samples of answers at each score level. This helps in verifying the correlation between Score values and actual answer quality. Based on these insights, we choose to retain answers with a Score of four or higher for subsequent processes. In the final phase, answers to the same question are ordered by their Score in descending order. Answers marked as ``accepted'' by the question questioner are prioritized by assigning them a theoretical Score of positive infinity, ensuring they appear first. For each question, we limit the dataset to the top five highest-scoring answers, thus optimizing the quality of data for model training purposes.

% For Zhihu, each training sample is structured with a question and its corresponding answer, organized as \texttt{Question: {question}\textbackslash n\textbackslash nAnswer: {answer}}. 
While Zhihu is recognized as one of the highest-quality QA platforms in China, the dataset still contains low-quality content such as advertisements, marketing materials, irrelevant or meaningless answers, and biased opinions. To enhance the data quality, we implement the following preprocessing steps. First, we assess user engagement by aggregating metrics such as upvotes, thanks, bookmarks, and followers for each user. Users who surpass a predefined threshold in these combined metrics are recognized as high-quality. Then, for each question, we retain answer that have obtained a substantial number of upvotes and authored by these high-quality users. Furthermore, we introduce a length limit, maintaining answers that are at least $200$ Chinese characters in length, or $100$ characters for responses from high-quality users. To further refine the dataset, we apply several heuristic rules aimed at eliminating promotional or irrelevant content. Specifically, answers containing the term ``editor'' more than twice are excluded, presuming them to be promotional. We also discard any sentences that begin with ``image source'' and apply additional filters for punctuation and formatting inconsistencies. These heuristic filters lead to the exclusion of approximately 2\% of the initial dataset.

% only the relatively highly upvoted answers from high-quality users, ensuring the answers are at least 100 Chinese characters long, or 200 characters for answers from any high-quality user. 

\subsection{Books}
% Lei Zhang
Books represent an invaluable data source for training LLMs, especially in fostering an understanding of long context dependency in natural language processing. High-quality books provide structured and detailed content that is crucial for enhancing the depth and scope of the model's comprehension capabilities. Particularly, textbooks have been proven to be exceptionally effective in improving LLMs' performance due to their rich, authoritative, and well-organized content~\cite{DBLP:journals/corr/abs-2306-11644,DBLP:journals/corr/abs-2309-05463}. Our pre-training dataset includes a diverse selection from the Books3 dataset, Project Gutenberg, CBook, Bestsellers, English textbooks, and Chinese textbooks, each offering unique advantages to the training process.

Books3 dataset is created by Shawn Presser and part of ``The Pile'' dataset~\cite{thepile}. It includes around $197,000$ books from Bibliotik in plain text format, covering a wide range of topics such as romance, fantasy, and science fiction. 

As one of the oldest digital libraries, Project Gutenberg offers a vast array of over $70,000$ free e-books in various languages. The library spans classic literature, fiction, non-fiction, and academic works. Although there is a ``frozen'' version of this corpus available as of 2018~\cite{Gerlach2018ASP}, we opt to collect the latest books up to May 2023 directly from the website, following deduplication and cleaning efforts, resulting in a total of $68,661$ English books.\footnote{\url{https://github.com/pgcorpus/gutenberg}}

CBook, made available by the Natural Language Processing Laboratory at Fudan University, includes about $150,000$ Chinese books covering diverse fields such as humanities, education, science, military, and politics. We use open sources for data acquisition and conduct a thorough cleaning operation to ensure quality.\footnote{\url{https://github.com/FudanNLPLAB/CBook-150K}}

Bestsellers comprises a selection of popular Chinese e-books, including textbooks and novels, sourced from Baidu Netdisk. These books are influential in their domains and contribute to the diversity of our training set.

English textbooks are manually collected from the Open Textbook Library.\footnote{\url{https://open.umn.edu/opentextbooks/}} They are downloaded in MOBI and EPUB formats, which are then converted into raw texts for training.

Chinese textbooks are acquired from the WanJuan corpus~\cite{wanjuan}. They have been preprocessed by the authors and are integrated into our training dataset to provide a rich source of educational content.

\paragraph{Preprocessing}
For CBook, we first convert all raw data~(containing MOBI and EPUB files) into plain text. For this conversion, we use \texttt{calibre} for MOBI files and the Python package \texttt{BeautifulSoup} for EPUB files.\footnote{\url{https://calibre-ebook.com}, \url{https://www.crummy.com/software/BeautifulSoup/bs4/doc/}}
After conversion, we apply heuristic rules to filter the text data to ensure relevance and readability: (1)~We exclude files containing fewer than $3,000$ characters, as these often lack sufficient content for meaningful training. (2)~Files where over $60\%$ of the lines contain fewer than six words are discarded due to their fragmented nature. (3)~Texts with less than $45\%$ Chinese characters are removed to maintain language consistency, as non-Chinese texts (\eg{} Korean or Japanese) may have been mistakenly included. Subsequent cleaning steps include the desensitization of sensitive information such as email addresses and phone numbers, and the removal of non-content elements like publishing details and navigational artifacts (\eg{} empty parentheses or brackets).

For Bestsellers, the raw data includes texts in TXT, EPUB, and MOBI formats. All files are converted to plain text and subsequently stored in JSONL format to streamline further processing. We apply heuristic filters to enhance data quality: (1) Files shorter than $170$ characters are removed to exclude incomplete or erroneously included texts. (2) We discard files where more than $29\%$ of lines are under six words long, as these are often poorly formatted or contain extraneous content. (3) Texts with less than $79\%$ Chinese characters are excluded to ensure the dataset primarily contains Chinese language material. Finally, we rigorously remove any remaining private information, such as publication numbers, website URLs, and contact details, to ensure privacy compliance and data integrity.

\subsection{News Articles}
% Qian Cao
News provides a stream of current events and real-time data that is crucial for training LLMs to be relevant and responsive to the latest global developments. By integrating news data from diverse sources, LLMs can better grasp the nuances of journalistic language, adapt to varying narrative styles, and improve their accuracy in information retrieval and generation tasks. In our dataset compilation, we include news from CC-news, RealNews~\cite{realnews}, and the news articles from China International Communications Group (CICG) to cover a wide range of topics and perspectives.

CC-news and RealNews are extensive corpora sourced from Common Crawl, specifically curated to include a wide array of news articles. We have accessed open-source versions of these datasets and have conducted thorough cleaning to ensure the removal of inappropriate content as delineated in Section~\ref{sec.data.cc} stage (1).\footnote{\url{https://huggingface.co/datasets/spacemanidol/cc-stories}, \url{https://github.com/rowanz/grover/tree/master/realnews}}

CICG, a key state-run foreign-language news and communication organization, provides rigorously vetted news content in English, Chinese, and other languages. This source is particularly valuable for obtaining reliable and official news narratives. We have access to news data spanning from April 2019 to May 2023 from proprietary sources, ensuring a rich dataset that reflects recent global events and trends.

\paragraph{Preprocessing}
For the CICG data, we initiate our preprocessing with a set of heuristic rules aimed at refining the quality of the data: (1) We discard files shorter than $170$ characters, as they often lack substantive content. (2) Articles where over $25\%$ of lines contain fewer than six words are excluded to eliminate fragments and poorly structured content. (3) We also filter out texts with less than $40\%$ Chinese characters to maintain consistency in language composition. Further cleaning involves the removal of non-essential elements such as: (1) The removal of headers and footers, including source attributions, publication dates, and editor names, to focus solely on the content. (2) The elimination of meaningless fragments, such as picture captions, formatting markers, and empty punctuation marks, to enhance the readability and relevance of the texts.

Through these preprocessing steps, we ensure that the news articles included in our training set are of the highest quality, free from extraneous text, and rich in valuable information, making them ideal for training LLMs.

% \subsubsection{Chinese Data}
\subsection{Legal Documents}
% Yutao Zhu
Legal judgment documents are also helpful for training LLMs due to their formal, structured nature and the logical complexity they embody. These documents encapsulate rigorous reasoning processes and legal terminology, making them beneficial for enhancing the analytical capabilities of LLMs. The precision and clarity required in legal language training help improve the model's ability to understand and generate text within specific, rule-based contexts, which is pivotal for applications in legal assistance, automated compliance checks, and advanced query-response systems in the legal domain.

We have curated a substantial corpus of legal judgment documents from China Judgments Online, accessed through a private and secure channel.\footnote{\url{https://wenshu.court.gov.cn/}} In order to adhere to privacy standards and legal requirements, sensitive information such as the names of the courts and judgment numbers are meticulously removed from the dataset. 

\subsection{Patents}
% Yutao Zhu
Patent applications are also useful for training LLMs due to their standardized format and formal, technical language. These documents are rich in specialized vocabulary and complex sentence structures, reflecting high levels of precision and clarity. Training LLMs on such data can significantly enhance their ability to parse and generate text within technical contexts. 

We incorporate patent applications into our pre-training, utilizing the WanJuan corpus~\cite{wanjuan} as our source. This corpus provides a vast and diverse collection of patent documents that have been pre-processed to meet training requirements. 

\subsection{Educational Assessments}
% Yutao Zhu
Educational assessments provide structured problem-solving environments that are vastly different from general text data, helping models learn to navigate and understand the specific formats and logical reasoning required in standardized testing. The inclusion of this type of data trains LLMs not only in content knowledge but also in the application of this knowledge within the constraints of a given question structure, which is crucial for achieving high performance on standardized assessments like MMLU~\cite{hendryckstest2021}, C-Eval~\cite{huang2023ceval}, and AGIEval~\cite{zhong2023agieval}.

To address this need, we incorporate a set of Chinese exam questions from the WanJuan corpus into our pre-training data. Although these questions constitute a relatively small portion of our dataset, they play a significant role in enhancing our LLM's ability to understand and correctly respond to the format and challenges presented by multi-choice tests. This approach effectively improves the model's performance on various comprehensive benchmarks, demonstrating the effectiveness of including targeted, format-specific training materials. More detailed analysis of the impact of this training is presented in Section~\ref{sec:exam}.

\subsection{Deduplication}
As demonstrated by existing studies~\cite{DBLP:conf/nips/XueFZZ023,d4}, data repetition may affect the performance of LLMs. Therefore, we perform deduplication within and across datasets. The deduplication tool is integrated in our YuLan-GARDEN tool~\cite{yulan-garden}.

\section{Evaluation Datasets}
\paragraph{TriviaQA} is originally designed as a reading comprehension dataset comprising over $650,000$ question-answer-evidence triples. Recently, this dataset has been repurposed for closed-book question answering tasks, where LLMs are prompted to answer questions without access to corresponding documents. For our evaluation, we utilize the test set of TriviaQA, which consists of approximately $17,000$ samples.

\paragraph{RACE} is a comprehensive reading comprehension dataset comprising over $28,000$ passages and nearly $100,000$ questions. It was collected from English examinations in China and is divided into two subsets: RACE-middle and RACE-high, tailored for middle school and high school students respectively.

\paragraph{CoQA} is a large-scale conversational question-answering dataset that features $8,000$ conversations and $127,000$ questions along with their corresponding answers.

\paragraph{CMRC2018} is a span-extraction dataset specifically designed for Chinese machine reading comprehension. It consists of approximately $20,000$ real questions that have been meticulously annotated on Wikipedia paragraphs by human experts.

\paragraph{C3} (Multiple-Choice Chinese machine reading Comprehension dataset) is a task that assesses machine reading comprehension skills utilizing prior knowledge, including linguistic, domain-specific, and general world knowledge. The dataset includes $13,369$ documents and $19,577$ multiple-choice free-form questions. It can be divided into two subsets: C3-dialog and C3-mixed, depending on the documents (dialogues or formally written mixed-genre texts).

\paragraph{GSM8K} (Grade School Math 8K) is a dataset of $8.5$K high-quality middle school math word problems, which involve performing a sequence of elementary calculations using basic arithmetic operations. 

\paragraph{AQuA-RAT} is a large-scale dataset consisting of approximately $100,000$ algebraic word problems. 

\paragraph{MMLU} (Measuring Massive Multitask Language Understanding) consists of multiple choice questions split into four domains: STEM, social sciences, humanities, and others. 

\paragraph{C-Eval} is a comprehensive Chinese evaluation suite for foundation models, which consists of $13,948$ multi-choice questions spanning $52$ diverse disciplines and four difficulty levels (STEM, Social Science, Humanity, and Other). 

\paragraph{Gaokao} (Chinese College Entrance Exam) is a subset in the AGIEval~\cite{zhong2023agieval}. The Gaokao dataset includes about $2,000$ examples from $8$ subjects: history, math, English, Chinese, geography, biology, chemistry, and physics.

\end{document}